\PassOptionsToPackage{dvipsnames}{xcolor}
\documentclass[11pt,letterpaper]{our}

\usepackage[all]{hypcap}
\usepackage{algorithm}
\usepackage{algorithmicx}
\usepackage{algpseudocode}
\usepackage{graphicx}
\usepackage{subcaption}
\usepackage{booktabs}
\usepackage{float}
\usepackage{bm}
\usepackage{bxcoloremoji}
\usepackage{mathtools}
\usepackage{mathrsfs}
\usepackage{nicefrac}
\usepackage{dsfont}
\usepackage{fdsymbol}
\usepackage{wrapfig}
\usepackage{lipsum}
\usepackage{stackengine}
\usepackage{adjustbox}
\usepackage{multicol}
\usepackage{multirow}
\usepackage{gradient-text}
\usepackage{rotating}
\usepackage[capitalise]{cleveref}
\usepackage{listings}

\PassOptionsToPackage{comma,numbers,compress}{natbib}
\usepackage[comma,authoryear,compress]{natbib}
\usepackage[font=small,labelfont=bf]{caption}

\definecolor{lightblue}{rgb}{0.22,0.45,0.70}
\definecolor{YaleBlue}{rgb}{0.06,0.30,0.56}

\newtcolorbox{AIbox}[2][]{aibox,title=#2,#1}

\usepackage{amsmath,amsfonts,bm}

\def\eqref#1{equation~\ref{#1}}

\def\1{\bm{1}}

\DeclareMathAlphabet{\mathsfit}{\encodingdefault}{\sfdefault}{m}{sl}
\SetMathAlphabet{\mathsfit}{bold}{\encodingdefault}{\sfdefault}{bx}{n}

\definecolor{darkergreen}{RGB}{50,160,50}

\newcommand{\euclid}{\gradientRGB{Euclid}{65,105,225}{128,0,128}} 
\newcommand{\github}{\raisebox{-1.5pt}{\includegraphics[height=1.05em]{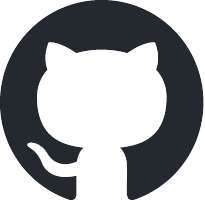}}}

\definecolor{myred}{HTML}{FF8577}
\definecolor{mygreen}{HTML}{0FA958}
\definecolor{myblue}{HTML}{1982C4}
\definecolor{codegreen}{rgb}{0,0.5,0}
\definecolor{codegray}{rgb}{0.5,0.5,0.5}
\definecolor{codepurple}{rgb}{0.07,0,0.53}
\definecolor{codered}{RGB}{189,41,0}
\definecolor{codecomment}{RGB}{153,153,153}
\definecolor{backcolour}{rgb}{0.96,0.96,0.96}
\definecolor{royalblue}{rgb}{0.0, 0.14, 0.4}
\definecolor{egyptianblue}{rgb}{0.06, 0.2, 0.65}
\definecolor{royalazure}{rgb}{0.0, 0.22, 0.66}
\definecolor{portlandorange}{rgb}{1.0, 0.35, 0.21}
\definecolor{sienna}{RGB}{183,105,68}
\definecolor{saddlebrown}{RGB}{139,69,19}
\definecolor{mediumbrown}{RGB}{83,41,11}
\definecolor{darkbrown}{RGB}{58,28,7}

\hypersetup{
    colorlinks=true,
    linkcolor=sienna,
    urlcolor=royalblue,
    citecolor=royalblue,
}

\usepackage{hyperref}
\hypersetup{
    colorlinks = true,
    citecolor = {YaleBlue},
}

\title{\euclid: Supercharging Multimodal LLMs with Synthetic High-Fidelity Visual Descriptions}
\author{Jiarui Zhang$^{\diamondsuit}$, Ollie Liu$^{\diamondsuit}$, Tianyu Yu$^{\spadesuit}$, Jinyi Hu$^{\spadesuit}$, Willie Neiswanger$^{\diamondsuit}$\\
$^{\diamondsuit}$University of Southern California, $^{\spadesuit}$Tsinghua University}

\correspondingauthor{Jiarui Zhang, \href{mailto:jzhang37@usc.edu}{jzhang37@usc.edu}; Willie Neiswanger, \href{mailto:neiswang@usc.edu}{neiswang@usc.edu}}

\begin{document}

\begin{abstract}
Multimodal large language models (MLLMs) have made rapid progress in recent years, yet continue to struggle with \emph{low-level visual perception} (LLVP)---particularly the ability to accurately describe the geometric details of an image. This capability is crucial for applications in areas such as robotics, medical image analysis, and manufacturing. 
In this paper, we first introduce \emph{Geoperception}, a benchmark designed to evaluate an MLLM’s ability to accurately transcribe 2D geometric information from an image. Using this benchmark, we demonstrate the limitations of leading MLLMs, and then conduct a comprehensive empirical study to explore strategies for improving their performance on geometric tasks. Our findings highlight the benefits of certain model architectures, training techniques, and data strategies, including the use of high-fidelity synthetic data and multi-stage training with a data curriculum. Notably, we find that a data curriculum enables models to learn challenging geometry understanding tasks which they fail to learn from scratch. Leveraging these insights, we develop \emph{Euclid}, a family of models specifically optimized for strong low-level geometric perception. Although purely trained on synthetic multimodal data, Euclid shows strong generalization ability to novel geometry shapes. For instance, Euclid outperforms the best closed-source model, Gemini-1.5-Pro, by up to 58.56\% on certain Geoperception benchmark tasks and 10.65\% on average across all tasks.

\vspace{3mm}
\coloremojicode{1F310} \textbf{Website}: \href{https://euclid-multimodal.github.io}{euclid-multimodal.github.io}

\coloremojicode{1F917} \textbf{Model Weights \& Datasets}: \href{https://huggingface.co/euclid-multimodal}{huggingface.co/euclid-multimodal}

\github{} \textbf{Code Repository}: \href{https://github.com/euclid-multimodal/Euclid}{github.com/euclid-multimodal/Euclid}
\end{abstract}
\maketitle

\vspace{2mm}
\section{Introduction}
\label{sec:intro}

Multimodal large language models (MLLMs) have rapidly progressed in recent years, demonstrating remarkable potential in understanding and reasoning about the visual world through the powerful capabilities of large language models (LLMs)~\citep{llava,llava15,gpt4,gemini,viscpm,cambrian,qwen2vl}.
These models have showcased strong performance in tasks such as visual question answering (VQA)~\citep{vqav2}, image captioning~\citep{coco}, and multimodal reasoning~\citep{mmbench}.
As one recent example, LLaVA-NeXT-34B~\citep{llavanext} achieves an impressive 83.7\% accuracy on the VQAv2 benchmark~\citep{vqav2}, a comprehensive benchmark on natural image question answering.

While MLLMs achieve impressive results on tasks like VQA, their performance relies on high-level semantic extraction~\citep{eyeswideshut}; in contrast, they often fall short on \emph{low-level visual perception} (LLVP)---i.e., the ability to accurately describe the geometric details of an image, such as the
points, lines, angles, shapes, and spatial relationships among its constituent objects.
This limitation becomes especially apparent in tasks requiring precise descriptions, such as mathematical visual problem solving~\citep{mathverse,mathvista}, scientific visual understanding~\citep{mmmu, isobench}, abstract visual reasoning~\citep{marvel,curiouscase}, and even simple visual comprehension~\citep{blind,svg}.
For example, when interpreting a graph diagram, precise recognition of edges is essential for extracting reliable information, and in geometry problem-solving, accurate identification of relationships between line segments and points is fundamental~\citep{isobench}.
Beyond abstract tasks, LLVP is also vital in real-world applications, including spatial understanding for robotics, medical image analysis for accurate diagnosis, quality control in manufacturing to detect subtle defects, autonomous driving systems that rely on exact object localization or distance estimation, and augmented reality applications that demand precise overlay of virtual objects onto the real world.

In this paper, we aim to study the challenges of LLVP in MLLMs, take steps to understand the root cause of their performance, and improve the models' capabilities in this area.
We begin by developing a benchmark dataset specifically designed to evaluate precise geometric perception, which we call \textit{Geoperception}.
As a focused test bed, this benchmark targets 2D geometry tasks.
Using this benchmark, we demonstrate the limitations of leading closed and open MLLMs, followed by a comprehensive empirical study to explore strategies for significantly improving MLLM's performance on geometric perception tasks.
Our findings show the benefits of key factors such as model architecture, training techniques, and data strategies, including the use of synthetic data and multi-stage training with a data curriculum.
Notably, we find that a data curriculum enables models to learn challenging geometry LLVP tasks, which they fail to learn from scratch, even when trained on a very large dataset.
Using these lessons learned, we then train a family of models---using a carefully designed curriculum of synthetic data---that are specifically optimized for strong LLVP, which we call \euclid.
We evaluate this family of models, and show that it excels on a variety of low-level geometric perception tasks.

\noindent Our main technical contributions are as follows:
\begin{itemize}[leftmargin=*,itemsep=2mm, topsep=0em]
    \item \textbf{Geoperception Benchmark:} We introduce a new benchmark dataset, \emph{Geoperception}, derived from the Geometry-3K corpus~\citep{geometry3k}, specifically designed to evaluate MLLMs' ability to accurately perceive surface-level geometric information without requiring complex inference or reasoning. Our benchmark reveals shortcomings in precise geometric perception across all leading vision-language MLLMs, both closed and open-source.
    \item \textbf{Empirical Study and Synthetic Data Engine:} To investigate the root cause of this performance, we conduct a detailed empirical exploration of MLLM architecture and training strategies. To aid in our investigation, we develop a synthetic data engine capable of generating high-fidelity visual descriptions of geometric elements. This study leads to key insights, such as the importance of certain architectural choices and the use of curriculum-based, multi-stage training with progressively more complex visual descriptions for improving low-level visual perception. 
    \item \textbf{Euclid Model Family:} Leveraging the insights from our exploration and our synthetic data engine, we train \euclid, a series of MLLMs tailored for high-quality geometric LLVP. Although purely trained on synthetic multimodal data with simple geometry shapes, \euclid{} generalizes strongly to the real-world geometry images from Geoperception benchmark, for instance, outperforming the best closed-source model, Gemini-1.5-Pro, by up to 58.56\% on certain benchmark tasks and 10.65\% across the tasks.
\end{itemize}
\section{Background and Related Work}

We provide an overview of prior efforts that assess and improve low-level perception and geometric reasoning in MLLMs, and highlight our contributions in data synthesis, evaluation, and training.

\paragraph{Vision-Language MLLMs.} While recent iterations of LLMs feature a standardized model architecture and pretraining recipe, MLLMs still often differ in design choices for infusing visual inputs. One popular design is to align \textit{continuous} visual features with the embedding space of a backbone LLM~\citep{llava15,llavanext,llama3,mm1,cambrian,paligemma,pixtral,qwen2vl}; another approach involves \textit{tokenizing} visual inputs to be trained jointly with language tokens~\citep{gemini,chameleon}. These modules are often infused with a decoder-only LLM, but others have explored encoder-decoder architectures to integrate a more varied collection of modalities~\citep{flamingo,4m,reka,4m21}. Our study focuses on \textit{decoder} MLLMs with a \textit{continuous} visual encoder, and we carry out an empirical study to explore the effect of synthetic dataset mixture, training recipe, and encoder design~\citep{convnext,clip,siglip,dinov2}.

\paragraph{Geometry-Oriented MLLMs.} At the core of these choices is the hardness in designing a module adept in general visual reasoning~\citep{mm1,cambrian}. In this work, we explore the optimal design of MLLMs specialized in low-level visual perception, a crucial aspect for (among other applications) multimodal mathematical understanding~\citep{mathvista,mathverse}.
This paper supplements prior efforts in improving mathematical reasoning~\citep{gllava,mavis,mathpuma,eaglemath,multimath,mathllava} with a detailed study on the effect of dataset mixture, curriculum, and visual encoder, to reach a recipe that elicits strong performance on geometric tasks~\citep{geomverse} that require low-level perception.

\paragraph{Evaluating LLVP.} Many benchmarks \citep{blind} have reported that frontier-class MLLMs struggle with visual perception tasks, which are prerequisites for applications that emphasize low-level geometric perception~\citep{spatialvlm,blink}, including mathematical~\citep{mmmu,mathvista,mathverse,marvel} and spatial reasoning~\citep{spatialvlm,tldr}. These findings collectively identify that MLLMs exhibit a language prior~\citep{languageprior}---a preference of textual inputs over visual inputs---leading to a performance gap between modalities~\citep{svg,mathverse,isobench}. Meanwhile, there lacks a high-quality benchmark that evaluates low-level geometric perception in MLLMs, and the Geoperception benchmark represents a first effort to narrow this gap. This type of efforts have led to significant improvements in certain capabilities of MLLMs, such as compositionality of objects~\citep{clip-bow,kong2023interpretable}.

\paragraph{Improving LLVP.} Many prior works study \textit{data-driven} approaches to improve low-level perception skills. For example,~\citet{gllava,eaglemath,mathpuma} employ a standardized supervised finetuning recipe, and optionally adjust the training data mixture. This type of training data is often synthesized from text-only math problems~\citep{geometry3k,alphageometry} or via rule-based systems~\citep{geomverse}. In parallel,~\cite{convnetvit,eagle,eyeswideshut} have explored the design space of visual encoders for general-purpose vision-language reasoning. We identify best practices over the union of these design spaces, and then train small MLLMs with strong performance in low-level perception tasks.

Lastly, several works \citep{toolformer,vipergpt,visualsketchpad} have opted to augment an MLLM with external APIs that process low-level features with specialized vision modules, such as object detection~\citep{yolo}, segmentation~\citep{sam}, and depth estimation~\citep{depthanything}. While these agentic frameworks \citep{autogen} present a promising alternative that directly addresses the shortcomings of visual encoders, they are limited by their scalability to novel use cases, and may be insufficient for precise tool routing that requires low-level perception as a primer~\citep{engineering,cad,cephalo}.
\section{Geoperception Benchmark}
\label{sec:dataset}

Recently, there has been a growing number of multimodal benchmarks across diverse domains beyond natural image understanding, including mathematical reasoning~\citep{mathverse,mathvista} and abstract visual reasoning~\citep{marvel,puzzlevqa}. Many of these prior works have realized the importance of accurate low-level visual perception. Specifically, Marvel~\citep{marvel} introduces perception questions for various abstract reasoning patterns, and finds that the main bottleneck of MLLMs' performance on abstract visual reasoning is that they fail to accurately transcribe visual information into concepts; Mathverse~\citep{mathverse} and IsoBench~\citep{isobench} both test MLLMs on equivalent question represented by language and visual modalities, respectively. Both works find that language-only input always outperforms vision-language input, and that the vision component of MLLMs always fails to utilize low-level visual features. VDLM~\citep{svg} transcribes raster images into vector graphics and uses LLMs to reason over the SVG code. They find that although SVG code is not straightforward to understand, using LLMs to reason over SVG is consistently more effective than directly using MLLMs on original raster images. Blind-test~\citep{blind} and BLINK~\citep{blink} also share similar findings with the works above.

\begin{figure*}[t]
  \centering
    \includegraphics[width=1\linewidth]{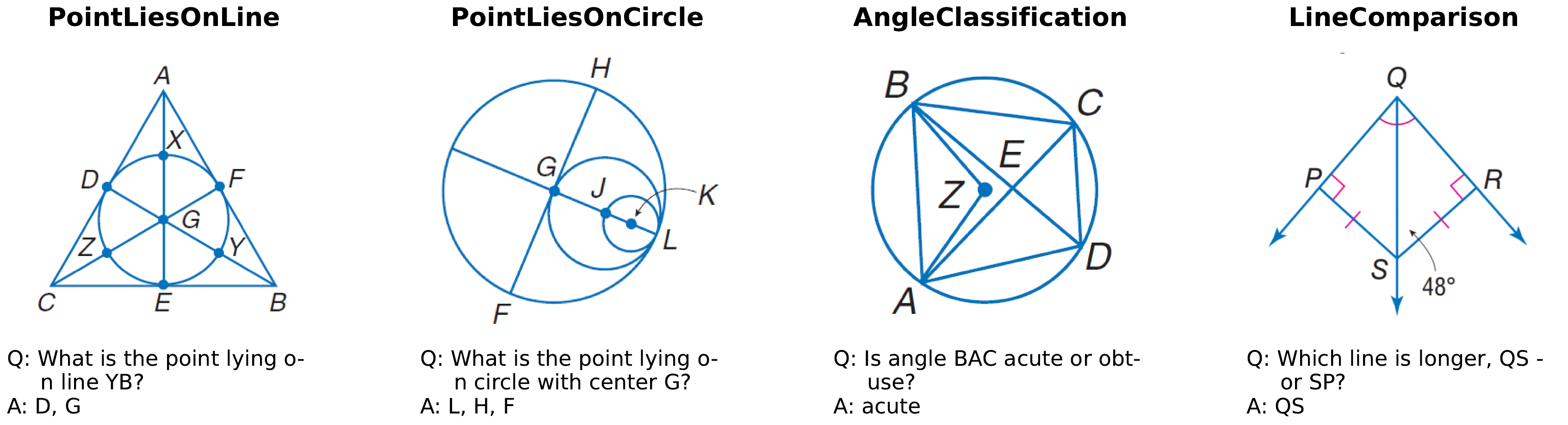}
    \caption{
    Four examples from our \textit{Geoperception} dataset. The questions are sourced from the Geometry-3K corpus~\citep{geometry3k}, which compiles problems from two widely-used high school textbooks. We perform filtering, validation, and generate question-and-answer text for each image.
    }
    \label{fig:geoperception_example_main}
\end{figure*}

\paragraph{A Benchmark for Geometric LLVP.}
Although such shortcomings of MLLMs are commonly recognized, there is a lack of comprehensive benchmark that purely focuses on these abilities of MLLMs. Our goal is to construct a benchmark focusing solely on the perception ability of MLLMs, which is also representative enough of real-world applications.
When humans perceive and memorize visual information, it is well-recognized that this procedure relies crucially on searching for the closest and simplest corresponding geometric shapes~\citep{humangeo}. We posit that geometric perception is a fundamental and broadly representative LLVP ability in many applications. Hence, we select geometry understanding as our domain of dataset construction.

\paragraph{Benchmark Tasks.}
Over two thousand years ago, Euclid introduced five axioms that underpin all further geometric reasoning. These axioms involve establishing and extending lines using points (Axioms 1 and 2), constructing circles from a point and a radius (Axiom 3), and defining perpendicularity (Axiom 4) and parallelism (Axiom 5). Additionally, Euclid provided common notions regarding the properties of equality.
To capture these aspects, we define five tasks in our Geoperception dataset: \texttt{PointLiesOnLine}, \texttt{PointLiesOnCircle}, \texttt{Parallel}, \texttt{Perpendicular} and \texttt{Equal}, and additionally define \texttt{AngleClassifi}-\texttt{cation} and \texttt{LineComparison} tasks to assess the model's understanding of angle and length measurements, resulting in a total of seven tasks.
 In geometric diagrams, perpendicularity, parallelism, and equality are often indicated by annotation symbols. Thus, we classify \texttt{Parallel}, \texttt{Perpendicular}, and \texttt{Equal} as annotated geometry understanding. Meanwhile, \texttt{PointLiesOnLine}, \texttt{PointLiesOnCircle}, \texttt{AngleClassification}, and \texttt{LineComparison} fall under primitive geometry shape understanding, which includes both logical (\texttt{PointLiesOnLine}, \texttt{PointLiesOnCircle}) and numerical (\texttt{AngleClassif}-\texttt{ication}, \texttt{LineComparison}) tasks.

\begin{wraptable}[13]{r}{7cm}
\vspace{-5mm}
\caption{Statistics of the seven tasks in our Geoperception dataset, including the number of questions and images.}
\vspace{2mm}
\centering
\begin{tabular}{ccc}
    \toprule
    Predicate                      & \# Q & \# I \\\midrule
    \texttt{PointLiesOnLine}       & 1901 & 924 \\
    \texttt{PointLiesOnCircle}     & 359  & 322 \\
    \texttt{Parallel}              & 106  & 101 \\
    \texttt{Perpendicular}         & 1266 & 456 \\
    \texttt{Equals}                & 4436 & 1202 \\
    \texttt{AngleClassification}   & 2193 & 1389 \\
    \texttt{LineComparison}      & 1394 & 1394 \\
    \bottomrule
\end{tabular}
\vspace{-20mm}
\label{tab:logic_form}
\end{wraptable}

\paragraph{Data Filtering.} Geoperception is sourced from the Geometry-3K~\citep{geometry3k} corpus, which offers precise logical forms for geometric diagrams, compiled from popular high-school textbooks. However, certain points in these logical forms are absent in the corresponding diagrams. To resolve this, we use GPT-4o-mini MLLM to confirm the presence of all points listed in the logical forms. This process filters the 3,002 diagrams to retain 1,584, where at least one logical form fully represents its points in the diagram. A random inspection of 100 annotations reveals only two errors, indicating high annotation accuracy.

\paragraph{Converting Logical Forms Into Questions.} We convert logical forms into question-and-answer pairs for each of the seven tasks in Geoperception. In the \texttt{Equals} task, for example, we directly convert the logical form (e.g., \texttt{Equals(LengthOf(Line(Q, T)), 86)}) into a question-answer pair (e.g., \texttt{Q: What is the length of line QT as annotated? A: 86}). For \texttt{PointLiesOnLine}, two points on the line are chosen to form the question, with the remaining points on the line as the answer. Similarly, for \texttt{PointLiesOnCircle}, we ask which points lie on the circle, using its center as the basis for the question. For \texttt{Parallel} and \texttt{Perpendicular}, we represent each line by two points and query which other lines are parallel or perpendicular to it. In \texttt{AngleClassification}, we ensure the queried angle is in the range of $[10, 80]\cup[100,170]$ degrees to avoid ambiguity. For \texttt{LineComparison}, we ensure that the shorter line is less than 70\% of the length of the longer line.
Since multiple equivalent questions can be generated for a single logical form (e.g., a line containing five points generates ${}^5P_2$ equivalent questions), we randomly select one to avoid redundancy. \cref{tab:logic_form} summarizes the question statistics for each task, as well as the number of images involved. Four examples from Geoperception are illustrated in~\cref{fig:geoperception_example_main}.

\paragraph{Evaluation Details.} We evaluate seven leading MLLMs, both open source and closed source. The open source models include Molmo-7B-D \citep{molmo}, Cambrian-1-8B \citep{cambrian}, Qwen2-VL-7B \citep{qwen2vl}, Llama-3.2-11B \citep{llama3}, and Pixtral-12B \citep{pixtral}. The closed-source models include GPT-4o-mini \citep{gpt4}, GPT-4o \citep{gpt4}, Claude-3.5-Sonnet \citep{claude3}, Gemini-1.5-flash \citep{gemini}, and Gemini-1.5-pro \citep{gemini}. Additionally, GPT-4o-mini without image input is used for generating the random baseline, employing the same textual instructions.
To prevent stretching, all images are padded to square dimensions before being fed into the models. During evaluation of a given question by an MLLM, let $G$ denote the ground truth set of answers, and let $P$ denote the predicted set of answers; then the evaluation score is defined as
\begin{equation}
\text{Evaluation score} =
\begin{cases}
\dfrac{|P|}{|G|} & \text{if } P \subseteq G, \\
0 & \text{otherwise}.
\end{cases}
\end{equation}

\paragraph{Current MLLMs struggle to perceive low-level geometry annotations and relationships.}
We show a comparison of all models on Geoperception in Table~\ref{tab:gpt_performance}.
Despite the simplicity of Geoperception for humans, it remains a considerable challenge for even the most advanced commercial MLLMs. Notably, all models fall short of achieving 30\% accuracy on the \texttt{PointLiesOnLine} task and do not outperform the text-only GPT-4o mini model in \texttt{AngleClassification} task. 
Closed source models generally outperform open source ones, with Gemini-1.5-pro attaining the highest average score of 56.98\%, followed by gemini-1.5-flash at 54.76\%. Among open source models, Pixtral-12B achieves the best performance with an overall score of 41.95\%. It is worth noting that Cambrian-1~\citep{cambrian}, which is reported to be trained on Geo-170K~\citep{gllava}, a geometry multimodal instruction tuning dataset built on the logical annotation of Geometry-3K, the same source with our Geoperception, still faces challenges in our Geoperception task, despite being trained on the dataset having the same images and augmented text annotations.
\vspace{2mm}

\begin{table}[h]
    \centering
    \caption{Performance (average evaluation score) of different models on Geoperception benchmark tasks. \texttt{POL}: PointLiesOnLine, \texttt{POC}: PointLiesOnCircle, \texttt{ALC}: AngleClassification, \texttt{LHC}: LineComparison, \texttt{PEP}: Perpendicular, \texttt{PRA}: Parallel, \texttt{EQL}: Equals. As the Random Baseline method, we use GPT-4o-mini, given the same textual instruction but without an image. The best model for each task is \textbf{bolded}.}
    \resizebox{\textwidth}{!}{
        \begin{tabular}{lcccccccc}
            \toprule
            & \multicolumn{2}{c}{Logical} & \multicolumn{2}{c}{Numerical} & \multicolumn{3}{c}{Annotations} & \\
            \cmidrule(lr){2-3} \cmidrule(lr){4-5} \cmidrule(lr){6-8}
            Model & \texttt{POL} & \texttt{POC} & \texttt{ALC} & \texttt{LHC} & \texttt{PEP} & \texttt{PRA} & \texttt{EQL} & Overall \\ \midrule
            Random Baseline                            &  1.35 &  2.63 & \textbf{59.92} & 51.36 &  0.23 &  0.00 &  0.02 & 16.50 \\ \midrule
            \multicolumn{9}{c}{ \textit{Open Source}}  \\
            Molmo-7B-D~\citep{molmo}             & 11.96 & 35.73 & 56.77 & 16.79 &  1.06 &  0.00 &  0.81 & 17.59 \\ 
            Llama-3.2-11B~\citep{llama3}         & 16.22 & 37.12 & 59.46 & 52.08 &  8.38 & 22.41 & 49.86 & 35.08 \\ 
            Qwen2-VL-7B~\citep{qwen2vl}          & 21.89 & 41.60 & 46.60 & 63.27 & 26.41 & 30.19 & 54.37 & 40.62 \\
            Cambrian-1-8B~\citep{cambrian}             & 15.14 & 28.68 & 58.05 & 61.48 & 22.96 & 30.74 & 31.04 & 35.44 \\
            Pixtral-12B~\citep{pixtral}          & 24.63 & 53.21 & 47.33 & 51.43 & 21.96 & 36.64 & 58.41 & 41.95 \\\midrule
            \multicolumn{9}{c}{ \textit{Closed Source}}  \\
            GPT-4o-mini~\citep{gpt4}             &  9.80 & 61.19 & 48.84 & 69.51 &  9.80 &  4.25 & 44.74 & 35.45 \\
            GPT-4o~\citep{gpt4}                  & 16.43 & \textbf{71.49} & 55.63 & 74.39 & 24.80 & 60.30 & 44.69 & 49.68 \\
            Claude 3.5 Sonnet~\citep{claude3}    & 25.44 & 68.34 & 42.95 & 70.73 & 21.41 & 63.92 & \textbf{66.34} & 51.30 \\
            Gemini-1.5-Flash~\citep{gemini}      & \textbf{29.30} & 67.75 & 49.89 & 76.69 & 29.98 & 63.44 & 66.28 & 54.76 \\
            Gemini-1.5-Pro~\citep{gemini}        & 24.42 & 69.80 & 57.96 & \textbf{79.05} & \textbf{38.81} & \textbf{76.65} & 52.15 & \textbf{56.98} \\
            \bottomrule
        \end{tabular}
    }
    \label{tab:gpt_performance}
\end{table}
\section{Empirical Study on MLLM Design Space}
\label{sec:empirical}

Although large-scale web-crawled image-text pairs cover a variety of domains, including geometry, the textual descriptions often lack the necessary specificity and depth.  To address this issue, current studies in this domain~\citep{gllava,mathllava,mavis} typically construct a geometry or mathematical domain dataset and apply the same training strategy used for general-purpose MLLMs. For example, Math-LLaVA~\citep{mathllava} and multi-math~\citep{multimath} rely on GPT-4V or GPT-4o's vision ability to generate most of the question and answer pairs and image captions, which is essentially model distillation. However, as evidenced by~\cref{tab:gpt_performance}, GPT-4o and Gemini-1.5-Pro often struggle to answer even basic perception questions, limiting the performance potential of resulting models. Furthermore, while works such as G-LLaVA~\citep{gllava}, MAVIS~\citep{mavis}, and Math-PUMA~\citep{mathpuma} utilize human crafted logical forms or synthetic multimodal data to ensure the reliability of textual annotations, they often conflate low-level perception with problem-solving, and train models to directly solve multimodal geometry problems, without verifying if the model's low-level perception abilities are sufficient. As an evidence, the best models in MAVIS~\citep{mavis} and Math-PUMA~\citep{mathpuma} evaluation results on Mathverse~\citep{mathverse} still have a substantial gap of 26.8\% and 28.7\% between text-dominant versions and vision-only versions of problems\footnote{In Mathverse, text-dominant is the version where the problem is mainly represented by text, while in the vision-only version an equivalent problem is represented purely by image.}, respectively. Furthermore, attempts to train MLLMs on low-level visual perception tasks~\citep{svg, blind} have also struggled to achieve satisfactory in-domain performance or generalize effectively. In this section, we aim to address these challenges.

\begin{wrapfigure}{r}{0.6\textwidth}
{
  \vspace{-5mm}
  \includegraphics[width=\linewidth]{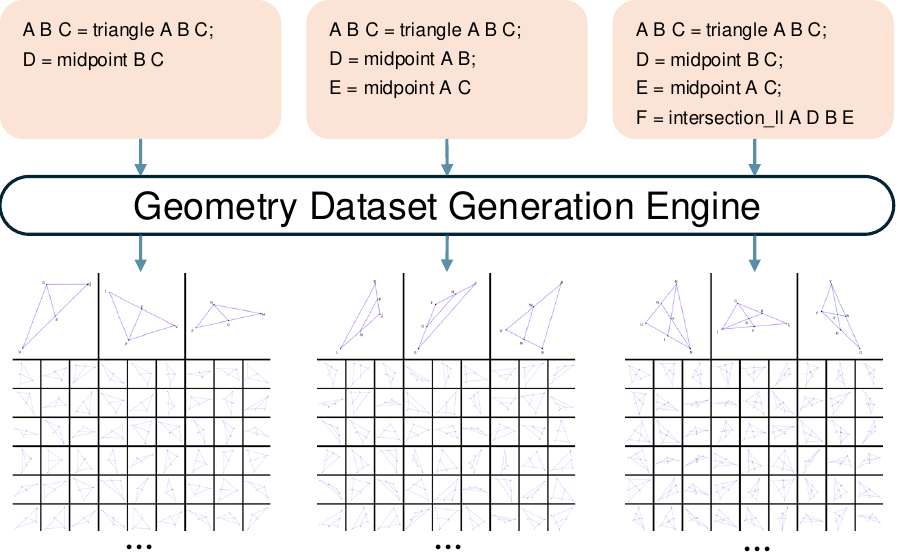}
  \vspace{-3mm}
  \caption{\small Three geometry logical shapes, of increasing complexity, used in our empirical study. Our geometry image generation engine is able to produce infinite visual instances for each of these logical shapes. All letters are randomly sampled from the alphabet and reassigned to each of the points before drawing.
  }
  \label{fig:three_shapes}
  \vspace{-8mm}
}
\end{wrapfigure}

We hypothesize the inability of today's MLLMs to effectively perceive basic geometric annotations and relationships stems from two factors: 1. The lack of high-fidelity geometric visual perception training data. 2. The problem of their model architectures and training strategy. Next, we will introduce our geometry dataset generation engine to overcome the lack of data, and then use generated dataset to study the optimal training strategy.

\paragraph{Geometry Image Generation Engine.} To provide sufficient high-fidelity training datasets, we develop a synthetic dataset generation engine to programmatically produce geometry shapes. 
Our geometry shape generation engine is built on AlphaGeometry~\citep{alphageometry}. 
Given an input formal language describing a geometry shape, the geometry engine will first check the validity of the geometry shape.
Then it will create numerical positions for all points following the restrictions given by the input. After the creation of all points, it will connect the line as specified in the input. To avoid inductive bias during training (e.g., point A is always on top of a triangle), letters are first picked from a letter pool (e.g., all 26 capital letters) and then randomly assigned to each point.
In addition to the original image generation engine, we introduce three visualization enhancements: (1) additional inputs to control the connections between points, number of letters in the letter pool, presence of each points, and annotation about length and angles; (2) increased randomness in creating numerical instances from conceptual shapes; and (3) adjustments to the canvas range to ensure visibility of all geometry components. Examples of our geometry generation engine, showing three geometries of increasing complexity, are shown in~\cref{fig:three_shapes}

\paragraph{Exploration of MLLM design space.}

With sufficient training dataset, we now revisit the MLLMs architectural and training design space. 
We choose 2 primitive geometry tasks from Geoperception as the test bed for the exploration: logical task, \texttt{PointLiesOnLine} and numerical task, \texttt{LineComparison}. For each task, we carefully create three tasks with incremental difficulty levels. We name them as difficulty level easy, medium and hard. Based on the insight from our preliminary experiments, to increase the difficulty levels, for \texttt{PointLiesOnLine}, we increase the complexity of geometry shapes as is shown in~\cref{fig:training_shapes}, for \texttt{LineComparison}, we increase the total number of letters in letter pool while mixing geometry shapes. During our preliminary experiments, we find that sometimes the model fails to converge due to instability. To this end, for all experiment moving forward, we run the training for three times and report the best run among them (i.e., having the lowest overall training loss or testing accuracy).

We start with a typical setting of MLLMs: CLIP-ViT-L/14~\citep{clip} as the visual encoder and a two layer MLP as multimodal connector and the latest Qwen-2.5 series~\citep{qwen2.5} as LLM. During training, we actively tune the MLP and LLM, while keeping visual encoder frozen. We use the mixture of three difficulty levels as the training set.

\begin{figure}[t]
  \centering
    \vspace{-1ex}
    \includegraphics[width=1\linewidth]{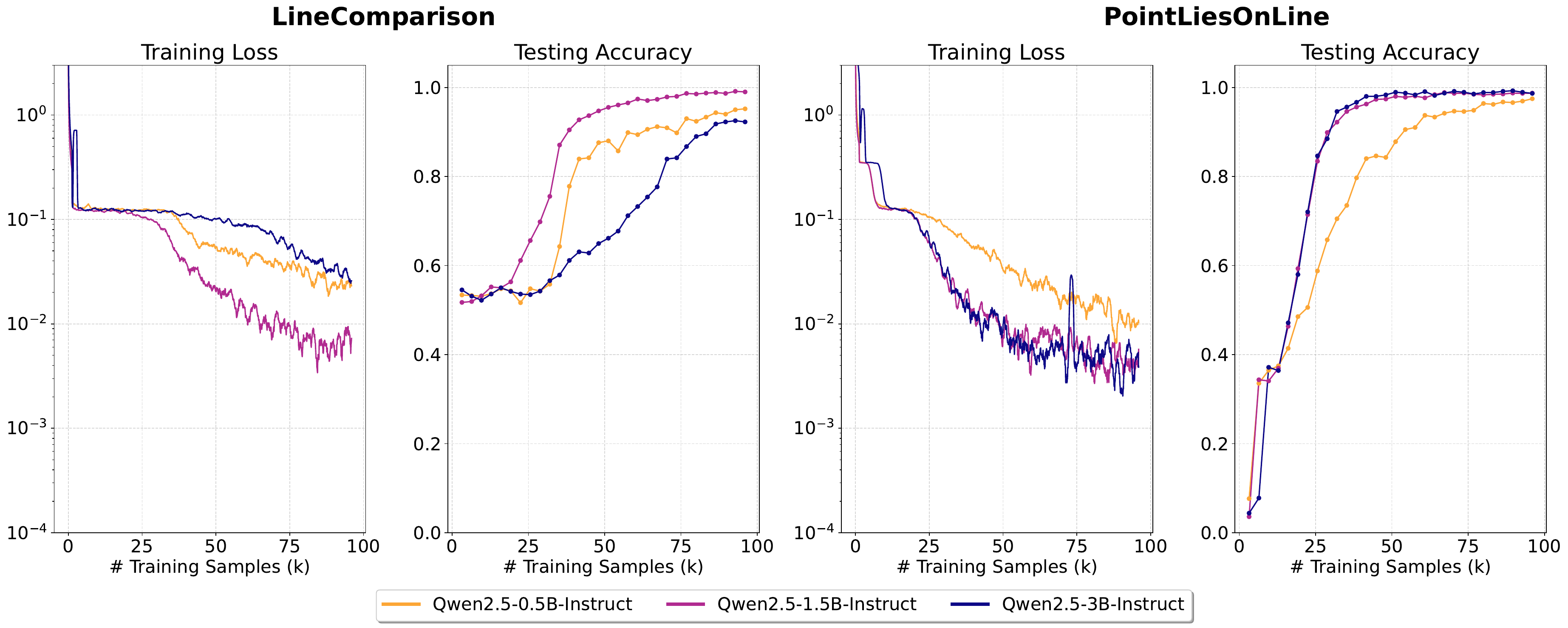}
    \vspace{-3ex}
    \caption{LLM size experiments. Training loss and testing accuracy curve comparing three choices of LLM size with a fixed visual encoder and multimodal connector.
    Training losses are window-smoothed using a window size of 10 for better visibility.}
    \label{fig:llm_size}
    \vspace{-1ex}
\end{figure}
\paragraph{Lesson 1: Under the same training dataset, scaling LLM sizes does not lead to better performance.}

It is commonly acknowledged that under the same training dataset, scaling up LLM can lead to better MLLM performance~\citep{llava15}. To this end, we first vary the sizes of LLMs, Qwen-2.5~\citep{qwen2.5} in a range of 0.5B, 1.5B, and 3B while keep other components consistent. The result is shown in~\cref{fig:llm_size}. For~\texttt{LineComparison}, Qwen-2.5-1.5B performs the best while Qwen-2.5-3B learns most slowly. For~\texttt{PointLiesOnLine}, Qwen-2.5-1.5B and Qwen-2.5-3B performs almost the same. Qwen-2.5-0.5B learns relatively slower, but still reach almost the same final performance with two other models. In conclusion, we do not observe an obvious trend that larger LLMs can learn such low-level visual perception task faster or better. Moving forward, we will use Qwen-2.5-1.5B to continue our exploration.

\begin{figure}[t]
  \centering
    \vspace{-1ex}
    \includegraphics[width=1\linewidth]{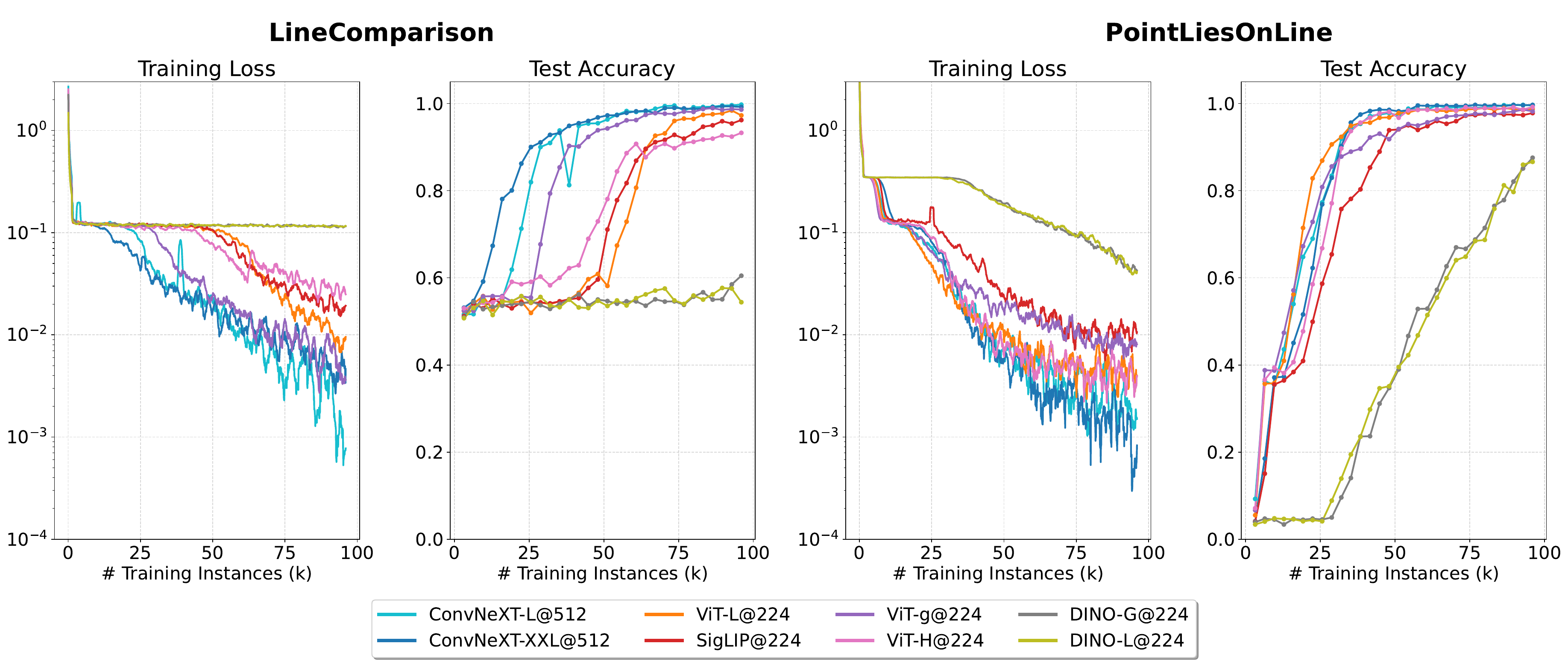}
    \vspace{-3ex}
    \caption{Vision encoder experiments. Training loss and testing accuracy (on a 1500 instances holdout test set) curve comparing eight visual encoders, with a fixed multimodal encoder and LLM. For a fair comparison, all visual encoder transcribe an image into 256 visual tokens. Training losses are window-smoothed using a window size of 10 for better visibility.}
    \label{fig:vienc_comparison}
    \vspace{-1ex}
\end{figure}
\begin{wraptable}[12]{r}{8cm}
\vspace{-3mm}
\caption{Summary of Visual Encoders}
\centering
\small
\begin{tabular}{lcc}
    \toprule
    Model & Params & Objective \\\midrule
    \href{https://huggingface.co/laion/CLIP-convnext_large_d_320.laion2B-s29B-b131K-ft-soup}{ConvNeXt Large@512} & 200M & CLIP \\
    \href{https://huggingface.co/laion/CLIP-convnext_xxlarge-laion2B-s34B-b82K-augreg-soup}{ConvNeXt XXLarge@512} & 847M & CLIP \\
    \href{https://huggingface.co/laion/CLIP-ViT-g-14-laion2B-s34B-b88K}{ViT-g/14@224} & 1.01B & CLIP \\
    \href{https://huggingface.co/laion/CLIP-ViT-H-14-laion2B-s32B-b79K}{ViT-H/14@224} & 632M & CLIP \\
    \href{https://huggingface.co/openai/clip-vit-large-patch14}{ViT-L/14@224} & 303M & CLIP \\
    \href{https://huggingface.co/google/siglip-so400m-patch14-224}{SigLIP@224 (ViT)} & 428M & CLIP-like \\
    \href{https://huggingface.co/facebook/dinov2-giant}{DINOv2 Giant@224 (ViT)} & 1.14B & Self-Sup \\
    \href{https://huggingface.co/facebook/dinov2-large}{DINOv2 Large@224 (ViT)} & 304M & Self-Sup \\
    \bottomrule
\end{tabular}
\label{tab:visual_encoders}
\end{wraptable}

\paragraph{Lesson 2: CNN architecture performs better than ViT.} 
We then study the choice of visual encoder architectures, including two families of architectures: Vision Transformer (ViT)~\citep{vit} and ConvNeXT~\citep{convnext}; as well as two visual representation learning objectives: language-supervised learning~\citep{clip} and self-supervised learning~\citep{dinov2}. We control the number of visual tokens to 256 for all of our vision encoders. The result is shown in ~\cref{fig:vienc_comparison}.
We find that ConvNeXt-XXLarge and ConvNeXt-Large consistently learns the geometric concept the fastest among all of the visual encoders. Notably, ConvNeXT-Large shows superior learning performance with the vision transformers which are 3-5 times larger. We hypothesize that CNN architecture extract visual features globally, effectively preserving low-level visual features. In contrast, ViT architectures split images into discrete patches, making it more challenging to retain the original low-level visual information.
Self-supervised learning (SSL) visual encoders, DINO-v2, struggles to learn the geometry concept; we hypothesis this is due to the weak vision-language representation in these models. Surprisingly, although the SigLIP-family is widely-recognized as a better visual encoder~\citep{cambrian}, we find that their performance in learning basic visual geometry attributes is limited.

\begin{figure*}[t]
  \centering
    \includegraphics[width=0.99\linewidth]{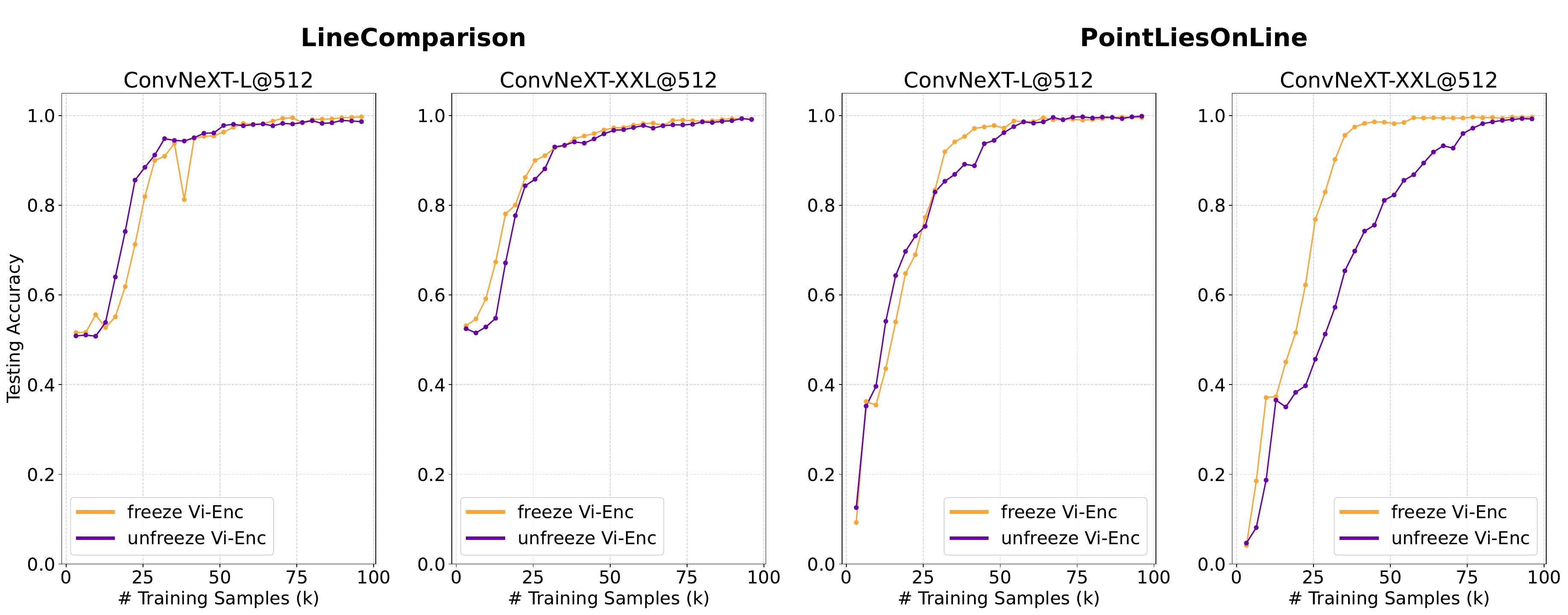}
    \caption{Tuning/freezing vision encoder experiments. Testing accuracy (on a 1500 instances holdout test set) curve comparing freezing versus tuning the visual encoder during training.}
    \label{fig:freeze_unfreeze}
\end{figure*}

\paragraph{Lesson 3: Tuning vision encoder does not provide significant help.}

We next study the effect of tuning versus freezing the visual encoder. In~\cref{fig:freeze_unfreeze}, we show the testing accuracy curves of tuning and freezing visual encoders. We find that compared with using a frozen encoder, tuning the visual encoder does not help the model learn low-level geometry relationships faster or better. In what follows, we will freeze the encoder for simplicity.

\paragraph{Lesson 4: Curriculum learning unleashes full potential.}
\label{curriculum}

\begin{wrapfigure}{r}{0.4\textwidth}
{
  \vspace{0mm}
  \includegraphics[width=\linewidth]{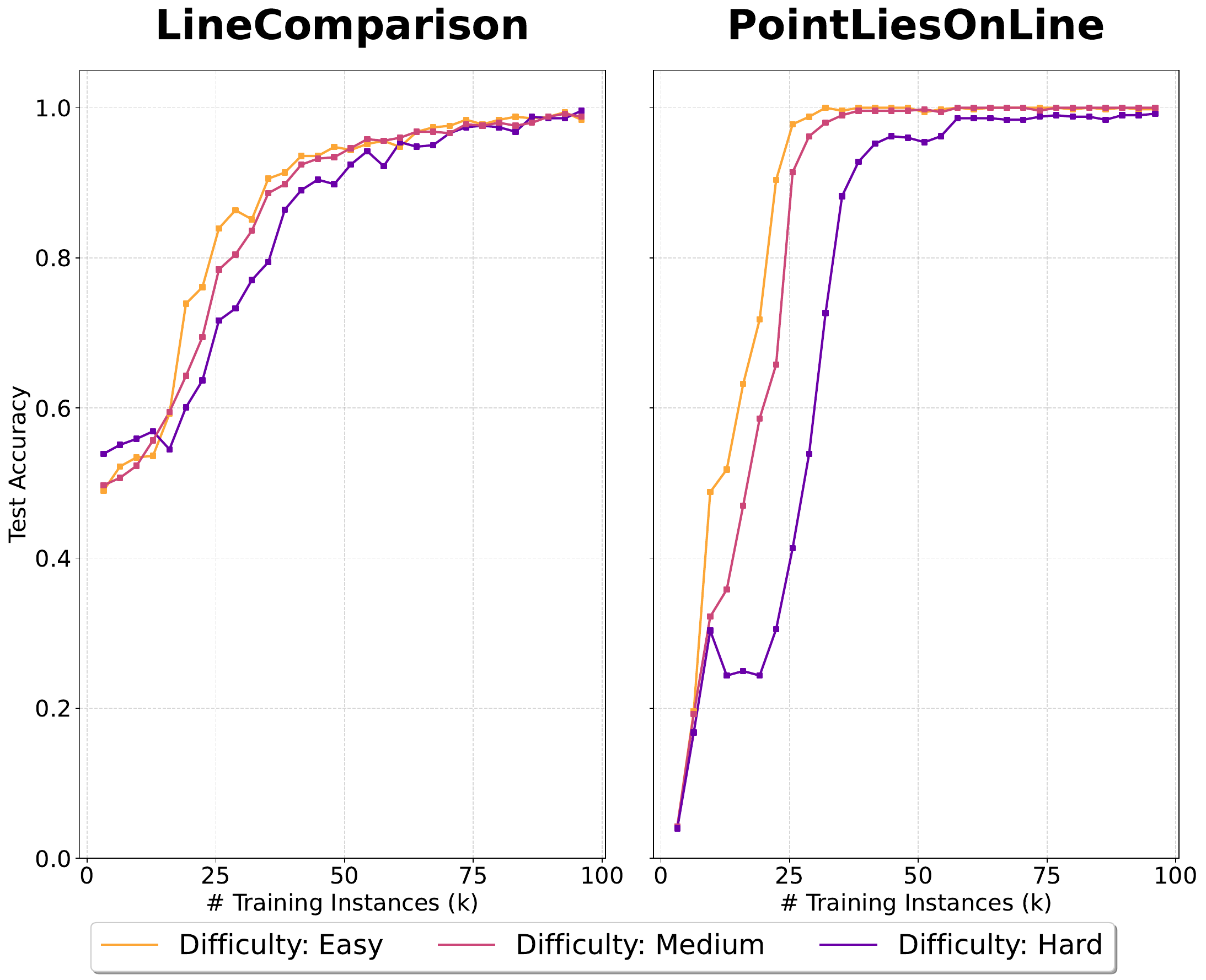}
  \caption{Separate testing accuracy curves on difficulty levels easy, medium, and hard, shown over the course of training on a mixture of all difficulty levels.}
  \label{fig:mixed_curriculum}
  \vspace{-1mm}
}
\end{wrapfigure}

Finally, we study training data composition. 
In our preliminary experiment~\cref{appendixfig:lesson_0}, we observe that the model fails to converge on difficulty level 3 of~\texttt{PointLiesOnLine} and difficulty level 2 and 3 of~\texttt{LineComparison}. However, when using mixed training set of all three difficulty levels, the model achieves convergence, despite using the same amount of data for each difficulty levels. We hypothesize that including easier levels aids the model in learning more complex levels. To test this hypothesis, we report the test accuracy for three difficulty levels separately during the mixed training of ConvNeXt-XXLarge, in~\cref{fig:mixed_curriculum}, on both tasks. We notice that the testing accuracy for easier tasks increase earlier and more quickly than difficulty tasks. In~\texttt{PointLiesOnLine} tasks, we notice an apparent plateau for hard level tasks until the model has trained on approximately 20K samples. During this period, the testing accuracy for easy and medium continue to increase. This suggests that learning easier shapes can significantly help the model tackle more challenging shapes, comparing with directly learning the challenging ones, this finding align with the principles of curriculum learning.

\begin{figure*}[t]
  \centering
    \includegraphics[width=0.97\linewidth]{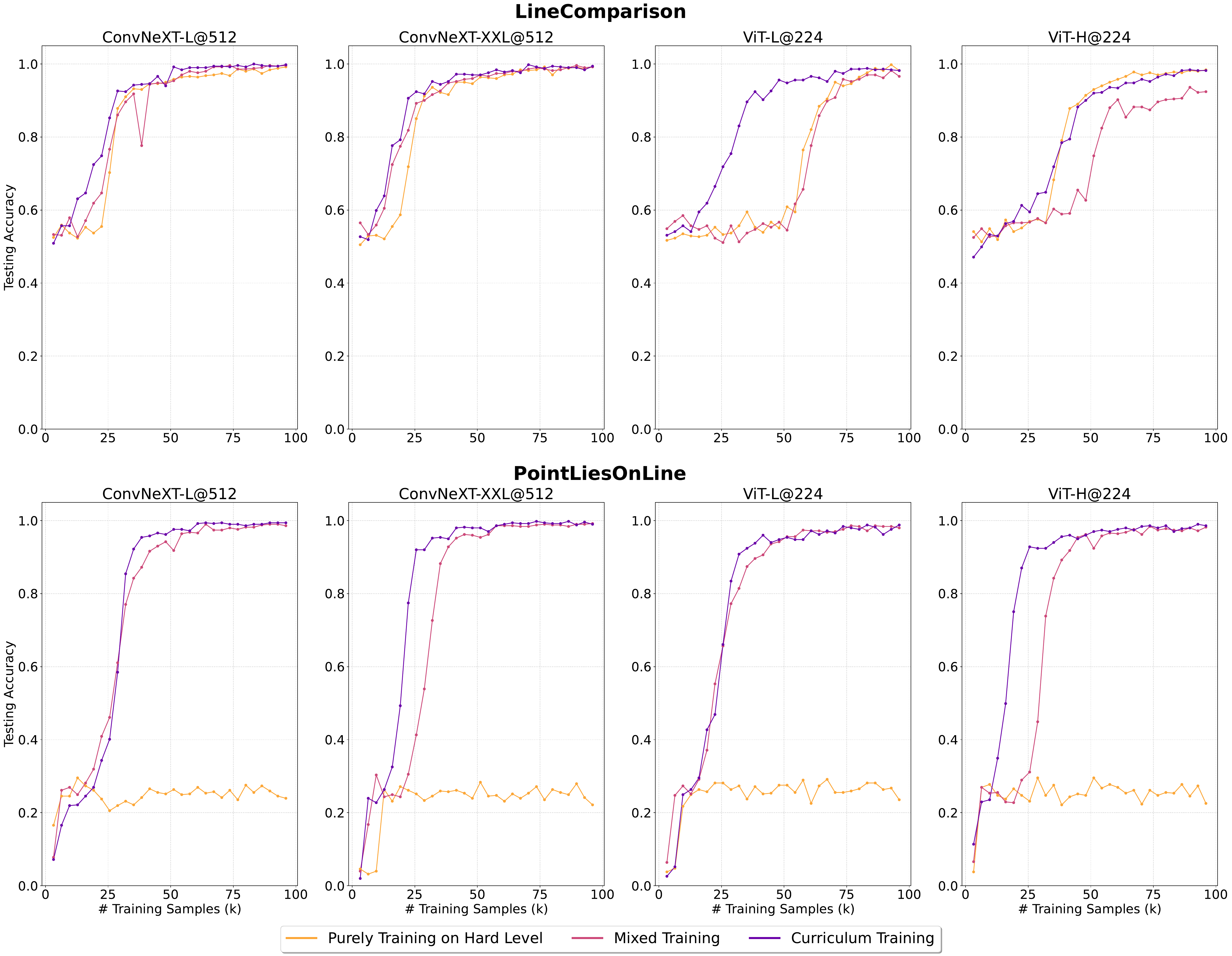}
    \vspace{0mm}
    \caption{Curriculum learning experiments. Test accuracy on difficulty level hard of three training strategies: purely training on difficulty level hard, mixed training of difficulty levels easy/medium/hard, and curriculum training.}
    \label{fig:curriculum_comparison}
    \vspace{2mm}
\end{figure*}

While mixed training enables effective spontaneous curriculum learning, we investigate whether a structured curriculum can further enhance model efficiency on challenging shapes. To this end, we train the model sequentially from simple to more complex shapes and compare testing accuracy just on hard level tasks. During training, we monitor the model's performance and dynamically adjust the distribution of training data (i.e., the curriculum stage) based on this performance. Specifically, the model starts by training on the easy level data. and is evaluated when it finishes a training round, using testing accuracy from the current level of data. Upon evaluation, if the model achieves an accuracy exceeding a predefined threshold \(\theta\), the framework advances the level to the next. Formally, the update rule for advancing stages is given by:
\begin{equation}
\text{if } \text{accuracy}_s > \theta \quad \Rightarrow \quad c \leftarrow c + 1.
\label{eq:update}
\end{equation}
The model is trained on a total of $M$ rounds and $K$ steps within each round. To avoid forgetting, we apply data smoothing at each stage. Specifically, we smooth our dataset distribution over all stages using an exponential attenuation function:
\begin{equation}
\text{ratio}_s = \exp\left(-\alpha \cdot |\text{stage}_s - c|\right),
\label{eq:ratio}
\end{equation}
where \(\alpha\) denotes the attenuation rate. \cref{eq:ratio} ensures that stages proximal to the current stage receive higher sampling probabilities.

We refer to this as our curriculum training strategy. Specifically, the accuracy threshold for advancing training stage $\theta$ is set to $0.99$. We train all the models for $M=30$ rounds, each round with $K = 50$ steps.
The results are shown in~\cref{fig:curriculum_comparison}. Firstly, we find that all of the models fail to converge when trained purely on hard level for~\texttt{PointLiesOnLine} task.
In contrast, the mixed training strategy shown by the red curve, consistently reaches faster convergence on hard level.
Curriculum training strategy, shown by the purple curve, proves more efficient than mixed training.
\section{\euclid: a Family of MLLMs for Geometric Visual Perception}

\label{sec:euclid}

In this section, we take all of the lessons we learned in the previous sections and train \euclid, a family of MLLMs specifically designed for strong geometric LLVP.

\paragraph{On-the-fly progressive training.}
We use the same strategy as the curriculum training in~\cref{curriculum}, but scale our training to all tasks in Geoperception. For each task, we create $N$ stages of training dataset shapes with progressively increasing geometric complexity.

\paragraph{Specifications.} 
For models, we select the best visual encoder architecture we found in our investigation, ConvNeXt, including ConvNeXt-Large@512 and ConvNeXt-XXLarge@512, and keep the same multimodal connector (2 layers MLP) and LLM (Qwen2.5-1.5B-instruct).
The accuracy threshold for advancing training stage $\theta$ is set to $0.99$. All models are trained on $N=3$ stages with manually curated geometry shapes and $M=50$ rounds with $K=500$ steps in each round, and the batch size is $64$ for each training step. The total training dataset volume for both of the models is 1.6M.

\begin{table}[h]
    \vspace{3mm}
    \centering
    \caption{Performance comparison between \euclid~and the best leading open source and closed source MLLMs on the seven tasks. Note that \euclid~is \textit{not} trained on any of the in-distribution data from the benchmark tasks below. The best model for each task is \textbf{bolded}.}
    \resizebox{\textwidth}{!}{
        \begin{tabular}{lcccccccc}
            \toprule
            & \multicolumn{2}{c}{Logical} & \multicolumn{2}{c}{Numerical} & \multicolumn{3}{c}{Annotations} & \\
            \cmidrule(lr){2-3} \cmidrule(lr){4-5} \cmidrule(lr){6-8}
            Model & \texttt{POL} & \texttt{POC} & \texttt{ALC} & \texttt{LHC} & \texttt{PEP} & \texttt{PRA} & \texttt{EQL} & Average \\ \midrule
            Random Baseline                      &  0.43 &  2.63 & 59.92 & 51.36 &  0.25 &  0.00 &  0.02 & 16.37 \\
            Pixtral-12B~\citep{pixtral}          & 24.63 & 53.21 & 47.33 & 51.43 & 21.96 & 36.64 & \textbf{58.41} & 41.95 \\
            Gemini-1.5-Pro~\citep{gemini}        & 24.42 & \textbf{69.80} & 57.96 & 79.05 & 38.81 & \textbf{76.65} & 52.15 & 56.98 \\\midrule
            \euclid-ConvNeXt-Large               & 80.54 & 57.76 & 86.37 & 88.24 & 42.23 & 64.94 & 34.45 & 64.93 \\
            \euclid-ConvNeXt-XXLarge             & \textbf{82.98} & 61.45 & \textbf{90.56} & \textbf{90.82} & \textbf{46.96} & 70.52 & 31.94 & \textbf{67.89} \\
            \bottomrule
        \end{tabular}
    }
    \label{tab:euclid_performance}
\end{table}

\paragraph{Evaluation results.}
The results are shown in~\cref{tab:euclid_performance}. Overall, although only trained on very simple synthetic geometry shapes, and using only a 1.5B language model, \euclid{} significantly outperforms current leading MLLMs in most of the tasks, showing strong generalization abilities on real-world geometry LLVP. Notably, in the \texttt{PointLiesOnLine} task, which is particularly challenging for existing MLLMs, \euclid{} achieves up to 82.98\% accuracy, more than three times the performance of Gemini-1.5-Pro. On all both numerical tasks, \texttt{LineComparison} and \texttt{AngleClassification}, \euclid{} keeps superior
performance. However, on three annotation tasks, \euclid{}'s performance is limited. We hypothesis this is due to the limited setting of our annotation types and styles, making the model hard to generalize to diverse human geometry annotations.

\paragraph{Error analysis.}
\begin{wrapfigure}{r}{0.28\textwidth}
{
  \vspace{-1mm}
  \centering
  \includegraphics[width=0.9\linewidth]{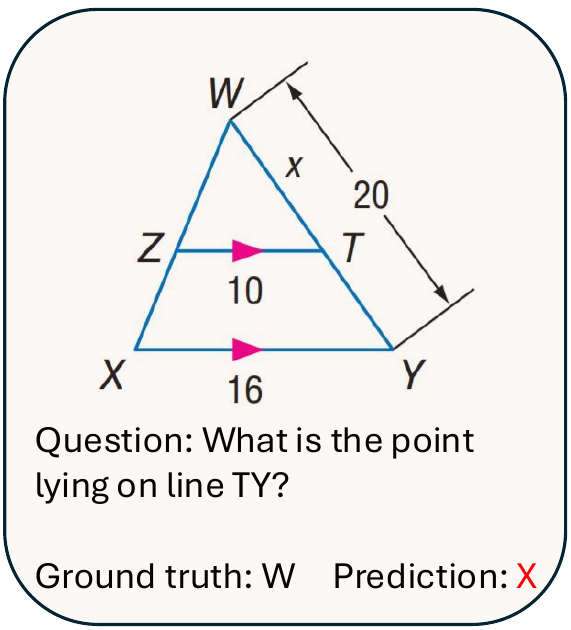}
  \caption{\small An error case where \euclid{} fails to predict the correct point on a line, potentially distracted by the annotation ``x''.}
  \label{fig:error_case}
  \vspace{-9mm}
}
\end{wrapfigure}
We take a deep look into \euclid{}'s prediction on Geoperception, we find that our model's performance is hindered when diagrams are heavily annotated. An example is shown in~\cref{fig:error_case}, where a line is annotated by ``x'', confusing the model from choosing the correct point. Incorporating training data with more diverse annotation types, geometry shape and can better distinguish different diagram annotation types could potentially help the model with such scenarios.
\vspace{-2mm}
\section{Conclusion and Future Work}
\label{sec:conclusion}
\vspace{-1mm}

In this work, we highlight the importance of accurate \emph{low-level visual perception}(LLVP) in MLLMs. To this end, we first introduce Geoperception, a large-scale multimodal benchmark focused exclusively on geometry-domain visual perception. We evaluate leading MLLMs on Geoperception, find that even top models such as Gemini-1.5-Pro struggle significantly it, although it is straightforward for humans. We then conduct an empirical study to explore the design space of MLLM training and architectures using the dataset generated by a geometric high-fidelity synthetic-data engine that we develop. Our study indicate that convolutional neural network visual encoders outperform vision transformers in our tasks; tuning the visual encoder generally enhances performance; and employing a curriculum-based training approach yields much more model potential than direct task training. Based on insights from this study, we develop \euclid, a model trained purely on high-fidelity synthetic generated data, which generalizes effectively to real-world geometric shape understanding tasks, surpassing the leading MLLMs by a substantial margin.

\textbf{Future work.} Our work examines the potential of using synthetic multimodal data to improve MLLM performance in low-level geometric perception tasks. However, there are still directions that remain under-explored: 
(1) Automatic curriculum learning. Incorporating a more diverse dataset, including varied geometric shapes and different domain dataset, introduces challenges in defining the learning order. Rule based definition and manual curation may become impractical, necessitating automated strategies like hard negative sampling to organize the curriculum based on training loss or testing accuracy. This approach could streamline the process, reduce human effort, provide more suitable and efficient curriculum learning orders.
(2) Using a more-diverse training dataset. Currently, the text portion of our synthetic multimodal training data uses a restricted set of templates, and the model trained on such templates could fail to generalize to other question types; it could therefore be beneficial to increase the diversity of our training images as well as the instruction-following formats. 
(3) Generalizing to other task domains. In this work, our study is focused on data from 2D geometry, as it provides a focused test bed of fundamental tasks. We believe the lessons we learn from this domain can be effectively generalized to a broader set of downstream domains that benefit from high-quality LLVP.

\vspace{-2mm}
\section*{Reproducibility Statement}
\vspace{-1mm}

In Section~\ref{sec:dataset}, we provide a comprehensive description of the procedure for generating the Geoperception benchmark. This includes employing GPT-4o-mini for dataset filtering and detailing the conversion of logical forms into questions and answers. Evaluation prompts for MLLMs on different types of Geoperception questions are presented in~\cref{fig:eval_prompts}. 
For model architecture exploration, we specify the visual encoders and provide corresponding Hugging Face links in~\cref{tab:visual_encoders}. Additionally, we outline the LLMs and multimodal connector architectures used. 
For our Euclid model, we include all geometry shape code used for training, along with demonstration diagrams and pseudo-code for generating training questions and answers.

\clearpage

\bibliography{cite}

\appendix
\newpage
\appendix

\section*{\center\LARGE Appendix}

\vspace{4mm}

\section{Geoperception Benchmark Details}
In Table~\ref{tab:logic_form_app}, we provide more details on the Geoperception benchmark, such as the number of logic forms present before and after filtering, the number of questions, and the number of images. \texttt{AngleClassification} and \texttt{LineComparison} are directly derived from points coordinates without filtering.

\begin{table}[h]
    \centering
    \begin{tabular}{ccccc}
        \toprule
        Predicate                  & \# LF Before Filter & \# LF After Filter & \# Q & \# I \\\midrule
        \texttt{PointLiesOnLine}   & 6988                & 2567               & 1901         & 924       \\
        \texttt{PointLiesOnCircle} & 1966                & 1240               & 359          & 322       \\
        \texttt{Parallel}          & 222                 & 123                & 106          & 101       \\
        \texttt{Perpendicular}     & 1111                & 680                & 1266         & 456       \\
        \texttt{Equals}            & 6434                & 4123               & 4436         & 1202      \\
        \bottomrule
    \end{tabular}
    \caption{Statistics of the five predicates in our Geoperception dataset. Including number of logic forms before filter, after filter and the number of questions and images.}
    \label{tab:logic_form_app}
\end{table}
\begin{figure*}[t]
  \centering
    \includegraphics[width=0.85\linewidth]{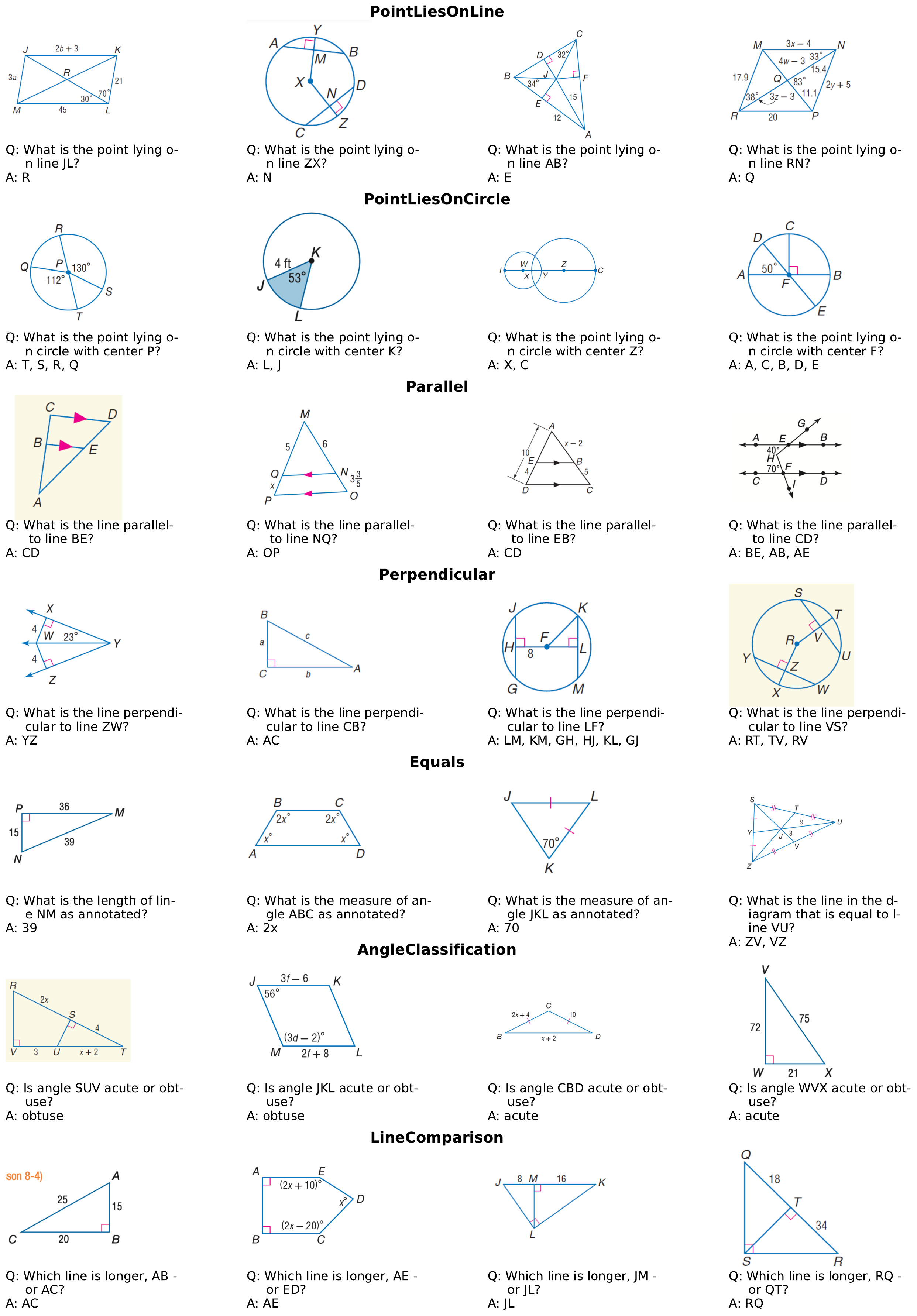}
    \caption{Examples of our \textit{Geoperception} dataset.}
    \label{fig:geoperception_example}
    \vspace{-3ex}
\end{figure*}

\clearpage
\section{Prompts for the Geoperception Dataset Evaluation}
\label{fig:eval_prompts}
\NewTColorBox{PolBox}{ s O {!htbp} }{%
  floatplacement={#2},
  IfBooleanTF={#1}{float*,width=\textwidth}{float},
  title=\textsc{Prompt Template for the PointLiesOnLine Task},
}

\begin{PolBox}
\vspace{1mm}
{\tt \small  
Answer me directly just with the all points lie on the line mentioned in the question (do not include the point mentioned in the question). 

Answer template: 

\hspace{4pt} (If only one point) The other point is: "your point". 

Or

\hspace{4pt} (if multiple points) The other points are: "your points".

For example: 

\hspace{4pt} The other point is: A

Or

\hspace{4pt} The other points are: A, B, C

}
\captionof{figure}{\textsc{Template for the PointLiesOnLine tasks}}\label{fig:pol-template}
\vspace{0.5mm}
\end{PolBox}
\NewTColorBox{PocBox}{ s O {!htbp} }{%
  floatplacement={#2},
  IfBooleanTF={#1}{float*,width=\textwidth}{float},
  title=\textsc{Prompt Template for the PointLiesOnCircle Task},
}

\begin{PocBox}
\vspace{1mm}
{\tt \small  








Answer me directly just with the all points lie on the circle mentioned in the question. 

Answer template: 

\hspace{4pt} (If only one point) The point is: "your point". 

Or 

\hspace{4pt} (If multiple points) The points are: "your points".

For example: 

\hspace{4pt} The point is: A

Or: 

\hspace{4pt} The points are: A, B, C

}
\captionof{figure}{\textsc{Template for the PointLiesOnCircle tasks}}\label{fig:poc-template}
\vspace{0.5mm}
\end{PocBox}
\NewTColorBox{ParallelBox}{ s O {!htbp} }{%
  floatplacement={#2},
  IfBooleanTF={#1}{float*,width=\textwidth}{float},
  title=\textsc{Prompt Template for the Parallel Task},
}

\begin{ParallelBox}
\vspace{1mm}
{\tt \small  








Answer me directly just with the all lines which are parallel to the line mentioned in the question (do not include the line mentioned in the question). 

Answer template: 

\hspace{4pt} (If only one line) The line is: "your line". 

Or 

\hspace{4pt} (If multiple lines) The lines are: "your lines".

For example: 

\hspace{4pt} The line is: BC

Or: 

\hspace{4pt} The lines are: BC, DE

}
\captionof{figure}{\textsc{Template for the Parallel tasks}}\label{fig:parallel-template}
\vspace{0.5mm}
\end{ParallelBox}
\NewTColorBox{PerpBox}{ s O {!htbp} }{%
  floatplacement={#2},
  IfBooleanTF={#1}{float*,width=\textwidth}{float},
  title=\textsc{Prompt Template for the Perpendicular Task},
}

\begin{PerpBox}
\vspace{1mm}
{\tt \small  








Answer me directly just with the all lines which are perpendicular to the line mentioned in the question (do not include the line mentioned in the question). 

Answer template: 

\hspace{4pt} (If only one line) The line is: "your line". 

Or 

\hspace{4pt} (If multiple lines) The lines are: "your lines".

For example: 

\hspace{4pt} The line is: BC

Or: 

\hspace{4pt} The lines are: BC, DE

}
\captionof{figure}{\textsc{Template for the Perpendicular tasks}}\label{fig:perp-template}
\vspace{0.5mm}
\end{PerpBox}
\NewTColorBox{EqBox}{ s O {!htbp} }{%
  floatplacement={#2},
  IfBooleanTF={#1}{float*,width=\textwidth}{float},
  title=\textsc{Prompt Template for the Equals Task},
}

\begin{EqBox}
\vspace{1mm}
{\tt \small  








Answer me directly just with the annotations presented on the image. 

Answer template:

\hspace{4pt} The annotation is: "your annotation".

For example:

\hspace{4pt} The annotation is: 2x+4

Or:

\hspace{4pt} The annotations is: 90

}
\captionof{figure}{\textsc{Template for the Equals tasks}}\label{fig:equals-template}
\vspace{0.5mm}
\end{EqBox}
\NewTColorBox{ClsBox}{ s O {!htbp} }{%
  floatplacement={#2},
  IfBooleanTF={#1}{float*,width=\textwidth}{float},
  title=\textsc{Prompt Template for the Angle Classification Task},
}

\begin{ClsBox}
\vspace{1mm}
{\tt \small  








Answer me directly just with the classification of the angle mentioned in the question. 

Answer template: 

\hspace{4pt} The angle is: "your angle".

For example: 

\hspace{4pt} The angle is: acute

Or:

\hspace{4pt} The angle is: obtuse

}
\captionof{figure}{\textsc{Template for the Angle Classification tasks}}\label{fig:cls-template}
\vspace{0.5mm}
\end{ClsBox}
\NewTColorBox{CmpBox}{ s O {!htbp} }{%
  floatplacement={#2},
  IfBooleanTF={#1}{float*,width=\textwidth}{float},
  title=\textsc{Prompt Template for the LineComparison Task},
}

\begin{CmpBox}
\vspace{1mm}
{\tt \small  








Answer me directly just with the longer line mentioned in the question. 

Answer template:

\hspace{4pt} The longer line is: "your line".

For example:

\hspace{4pt} The longer line is: BC

Or:

\hspace{4pt} The longer line is: DE
}
\captionof{figure}{\textsc{Template for the LineComparison tasks}}\label{fig:cmp-template}
\vspace{0.5mm}
\end{CmpBox}

\clearpage
\section{Details for Training Data Engine}

In this section, we provide all geometry shapes we use for \euclid{} training, including the pseudocode for generating text describing the geometry shapes and diagram examples.
\label{sec:question_generation}
\subsection{Pseudocode for Training Textual Dataset Synthesis}

\begin{algorithm}
\caption{Data Synthesis for the \textsc{PointLiesOnLine} Task}
\begin{algorithmic}[1]
\State \textbf{Input:} \texttt{data\_info}, \texttt{points\_set}
\State \textbf{Output:} \texttt{data}
\For{\texttt{points\_set} $\in$ \texttt{data\_info}}
    \For{\texttt{(A, B)} $\in$ \texttt{permutations(points\_set, 2)}}
        \State \texttt{all\_rest\_points $\gets$ [p for p in points\_set if p not in [A, B]]}
        \For{\texttt{rest\_points} $\in$ \texttt{permutations(all\_rest\_points)}}
            \State \texttt{verb\_agreement $\gets$ 'is' if len(rest\_points) == 1 else 'are'}
            \State \texttt{rest\_points $\gets$ [f"\{p\}" for p in rest\_points]}
            \State \texttt{rest\_points $\gets$ sorted(rest\_points)}
            \State \texttt{question $\gets$ 'What is the point lying on line ' + A + B + '?'}
            \State \texttt{answer $\gets$ 'The point lying on line ' + A + B + ' ' + verb\_agreement + ' ' + ', '.join(rest\_points)}
            \State \texttt{gt $\gets$ ''.join(rest\_points)}
            \State \texttt{data $\gets$ \{'question': question, 'answer': answer, 'gt': gt\}}
        \EndFor
    \EndFor
\EndFor
\end{algorithmic}
\end{algorithm}


\begin{algorithm}
\caption{Data Synthesis for the \textsc{PointLiesOnCircle} Task}
\begin{algorithmic}[1]
\State \textbf{Input:} \texttt{data\_info}
\State \textbf{Output:} \texttt{data}

\State \texttt{point\_set $\gets$ random.choice(list(data\_info.items()))}
\State \texttt{center\_point $\gets$ point\_set[0]}
\State \texttt{target\_points $\gets$ point\_set[1]}
\State \texttt{target\_points $\gets$ sorted(target\_points)}
\State \texttt{question $\gets$ 'What are the point lying on circle ' + center\_point + '?'}
\State \texttt{answer $\gets$ 'The point lying on circle ' + center\_point + ' are ' + ', '.join(target\_points)}
\State \texttt{gt $\gets$ ''.join(target\_points)}
\State \texttt{data $\gets$ \{'question': question, 'answer': answer, 'gt': gt\}}

\end{algorithmic}
\end{algorithm}
\clearpage

\begin{algorithm}
\caption{Data Synthesis for the \textsc{AngleClassification} Task}
\begin{algorithmic}[1]
\State \textbf{Input:} \texttt{data\_info}
\State \textbf{Output:} \texttt{data}
\State \texttt{angle $\gets$ data\_info}
\State \texttt{angle\_options $\gets$ [f'\{angle[1][0]\}\{angle[1][1]\}\{angle[1][2]\}', f'\{angle[1][2]\}\{angle[1][1]\}\{angle[1][0]\}']}
\State \texttt{angle\_letter $\gets$ random.choice(angle\_options)}
\State \texttt{angle\_class $\gets$ 'acute' if angle[0] < 90 else 'obtuse'}
\State \texttt{question $\gets$ 'Is angle ' + angle\_letter + ' acute or obtuse?'}
\State \texttt{answer $\gets$ 'Angle ' + angle\_letter + ' is ' + angle\_class}
\State \texttt{gt $\gets$ angle\_class}
\State \texttt{data $\gets$ \{'question': question, 'answer': answer, 'gt': gt\}}
\end{algorithmic}
\end{algorithm}


\begin{algorithm}
\caption{Data Synthesis for the \textsc{LineComparison} Task}
\begin{algorithmic}[1]
\State \textbf{Input:} \texttt{data\_info}
\State \textbf{Output:} \texttt{data}
\State \texttt{names $\gets$ [data\_info[0][1], data\_info[1][1]]}
\State \texttt{lengths $\gets$ [data\_info[0][0], data\_info[1][0]]}
\If{\texttt{lengths[0] > lengths[1]}}
    \State \texttt{longer\_name, shorter\_name $\gets$ names[0], names[1]}
\Else
    \State \texttt{longer\_name, shorter\_name $\gets$ names[1], names[0]}
\EndIf
\State \texttt{data $\gets$ [}
\State \ \ \ \ \texttt{\{ 'question': 'Which line is longer, ' + longer\_name + ' or ' + shorter\_name + '?',}
\State \ \ \ \ \texttt{\ \ 'answer': 'The longer line is ' + longer\_name,}
\State \ \ \ \ \texttt{\ \ 'gt': longer\_name \},}
\State \ \ \ \ \texttt{\{ 'question': 'Which line is longer, ' + shorter\_name + ' or ' + longer\_name + '?',}
\State \ \ \ \ \texttt{\ \ 'answer': 'The longer line is ' + longer\_name,}
\State \ \ \ \ \texttt{\ \ 'gt': longer\_name \}}
\State \texttt{]}
\end{algorithmic}
\end{algorithm}

\begin{algorithm}
\caption{Data Synthesis for the \textsc{Parallel} Task}
\begin{algorithmic}[1]
\State \textbf{Input:} \texttt{data\_info}
\State \textbf{Output:} \texttt{data}
\State \texttt{points\_set} $\gets$ \texttt{data\_info}
\For{\texttt{line\_points} $\in$ \texttt{points\_set}}
    \For{\texttt{(A, B)} $\in$ \texttt{permutations(line\_points, 2)}}
        \State \texttt{all\_rest\_lines $\gets$ [p for p in points\_set if p != line\_points]}
        \State \texttt{gts $\gets$ [`'.join(}
        \State \ \ \ \ \ \ \ \ \texttt{f`\{p\}' for line in all\_rest\_lines for p in line)}
        \State \texttt{]}
        \State \texttt{rest\_point\_pairs $\gets$ []}
        \For{\texttt{rest\_line} $\in$ \texttt{all\_rest\_lines}}
            \State \texttt{C, D $\gets$ random.sample(rest\_line, 2)}
            \State \texttt{rest\_point\_pairs.append([C, D])}
        \EndFor
        \State \texttt{all\_possible\_answer $\gets$ `, '.join(}
        \State \ \ \ \ \ \texttt{[f`\{C\}\{D\}' for C, D in rest\_point\_pairs]}
        \State \texttt{)}
        \State \texttt{verb\_agreement $\gets$ `is' if len(rest\_point\_pairs) == 1 else `are'}
        \State \texttt{question $\gets$ `What is the line parallel to line ' + A + B + `?'}
        \State \texttt{answer $\gets$ (}
        \State \ \ \ \ \ \texttt{`According to the diagram, the line parallel to ' + }
        \State \ \ \ \ \ \ \texttt{A + B + verb\_agreement + all\_possible\_answer}
        \State \texttt{)}
        \State \texttt{gt $\gets$ `, '.join(gts)}
        \State \texttt{data $\gets$ \{ }
        \State \ \ \ \ \ \texttt{`question': question, `answer': answer, `task': task, `gt': gt}
        \State \texttt{\}}
    \EndFor
\EndFor
\end{algorithmic}
\end{algorithm}

\begin{algorithm}
\caption{Data Synthesis for the \textsc{Perpendicular} Task}
\begin{algorithmic}[1]
\State \textbf{Input:} \texttt{data\_info}
\State \textbf{Output:} \texttt{data}
\State \texttt{source\_lines}, \texttt{target\_lines} $\gets$ \texttt{data\_info}
\State \texttt{all\_possible\_answer $\gets$ []}
\State \texttt{gts $\gets$ target\_lines}
\Comment{Randomly choose two points from each target line}
\For{\texttt{target\_line} $\in$ \texttt{target\_lines}}
    \State \texttt{C, D $\gets$ random.sample(target\_line, 2)}
    \State \texttt{all\_possible\_answer.append(f`\{C\}\{D\}')}
\EndFor
\State \texttt{verb\_agreement $\gets$ `is' if len(all\_possible\_answer) == 1 else `are'}
\For{\texttt{(A, B)} $\in$ \texttt{permutations(source\_line, 2)}}
    \State \texttt{question $\gets$ `What is the line perpendicular to line ' + A + B + `?'}
    \State \texttt{answer $\gets$ (}
    \State \ \ \ \ \ \texttt{`According to the diagram, the line perpendicular to ' + }
    \State \ \ \ \ \ \ \texttt{A + B + verb\_agreement + `, '.join(all\_possible\_answer}
    \State \texttt{)}
    \State \texttt{gt $\gets$ `, '.join(gts)}
    \State \texttt{data $\gets$ \{}
    \State \ \ \ \ \ \texttt{`question': question, ', `answer': answer, '`ask': task, `gt': gt}
    \State \texttt{\}}
\EndFor
\end{algorithmic}
\end{algorithm}

\begin{algorithm}
\caption{Data Synthesis for the \textsc{Equal} Task}
\begin{algorithmic}[1]
\State \textbf{Input:} \texttt{data\_info}
\State \textbf{Output:} \texttt{data}
\State \texttt{statement, content $\gets$ data\_info.split(`;')}
\If{\texttt{statement == `angles\_value'}}
    \State \texttt{angle\_letter, angle\_measure $\gets$ content.split(`=')}
    \State \texttt{angle\_letter $\gets$ random.choice([angle\_letter, angle\_letter[::-1]])}
    \State \texttt{question $\gets$ `What is the measure of angle ' + angle\_letter + ` as annotated?'}
    \State \texttt{answer $\gets$ `Angle ' + angle\_letter + ` is annotated as ' + angle\_measure}
    \State \texttt{gt $\gets$ angle\_measure}
\ElsIf{\texttt{statement == `segments\_value'}}
    \State \texttt{segment\_letter, segment\_length $\gets$ content.split(`=')}
    \State \texttt{segment\_letter $\gets$ random.choice([segment\_letter, segment\_letter[::-1]])}
    \State \texttt{question $\gets$ `What is the length of line ' + segment\_letter + ` as annotated?'}
    \State \texttt{answer $\gets$ `Line ' + segment\_letter + ` is annotated as ' + segment\_length}
    \State \texttt{gt $\gets$ segment\_length}
\ElsIf{\texttt{statement == `angles'}}
    \State \texttt{angle1, angle2 $\gets$ content.split(`=')}
    \State \texttt{angle1 $\gets$ random.choice([angle1, angle1[::-1]])}
    \State \texttt{angle2 $\gets$ random.choice([angle2, angle2[::-1]])}
    \State \texttt{query\_angle $\gets$ random.choice([angle1, angle2])}
    \State \texttt{answer\_angle $\gets$ angle2 if query\_angle == angle1 else angle1}
    \State \texttt{question $\gets$ `What is the angle in the diagram that is equal to angle ' + query\_angle}
    \State \texttt{answer $\gets$ `Angle ' + query\_angle + ` is equal to angle ' + answer\_angle}
    \State \texttt{gt $\gets$ answer\_angle}
\ElsIf{\texttt{statement == `segments'}}
    \State \texttt{segment1, segment2 $\gets$ content.split(`=')}
    \State \texttt{segment1 $\gets$ random.choice([segment1, segment1[::-1]])}
    \State \texttt{segment2 $\gets$ random.choice([segment2, segment2[::-1]])}
    \State \texttt{query\_segment $\gets$ random.choice([segment1, segment2])}
    \State \texttt{answer\_segment $\gets$ segment2 if query\_segment == segment1 else segment1}
    \State \texttt{question $\gets$ `What is the segment in the diagram that is equal to segment ' + query\_segment}
    \State \texttt{answer $\gets$ `Segment ' + query\_segment + ` is equal to segment ' + answer\_segment}
    \State \texttt{gt $\gets$ answer\_segment}
\EndIf
\State \texttt{data $\gets$ \{}
\State \ \ \ \ \ \texttt{`question': question, `answer': answer, `task': task, `gt': gt}
\State \texttt{\}}
\end{algorithmic}
\end{algorithm}

\clearpage
\subsection{Geometry Shapes Used for \euclid{} Training}

\NewTColorBox{GeoShapeGenBox}{ s O {!htbp} }{%
  floatplacement={#2},
  IfBooleanTF={#1}{float*,width=\textwidth}{float},
  title=\textsc{Geometry Shape Generation Code}
}

\begin{GeoShapeGenBox}
\vspace{1mm}
{\tt \tiny  

\textbf{PointLiesOnLine}

\hspace{4pt} (stage 1) A B C = triangle A B C; D = midpoint B C

\hspace{4pt} (stage 1) A B C = triangle A B C; D = midpoint B C; O = circle O A B C

\hspace{4pt} (stage 2) A B C = triangle A B C; D = midpoint A B; E = midpoint A C

\hspace{4pt} (stage 2) A B C = triangle A B C; D = midpoint A B; E = midpoint A C; O = circle O A B C

\hspace{4pt} (stage 3) A B C = triangle A B C; D = midpoint B C; E = midpoint A C; F = intersection\_ll A D B E

\hspace{4pt} (stage 3) A B C = triangle A B C; D = midpoint B C; E = midpoint A C; F = intersection\_ll A D B E; O = circle O A B C

\textbf{PointLiesOnCircle}

\hspace{4pt} (stage 1) A B = segment A B; C = on\_circle C A B

\hspace{4pt} (stage 1) A B = segment A B; C = on\_circle C A B; D = on\_circle D A B

\hspace{4pt} (stage 1) A B = segment A B; C = on\_circle C A B; D = on\_circle D A B; E = on\_circle E A B

\hspace{4pt} (stage 1) A B = segment A B; C = on\_circle C A B; D = on\_circle D A B; E = on\_circle E A B; F = on\_circle F A B

\hspace{4pt} (stage 1) A B = segment A B; C = on\_circle C A B; D = on\_circle D A B; E = on\_circle E A B; F = on\_circle F A B; G = on\_circle G A B

\hspace{4pt} (stage 2) A B = segment A B; C = on\_circle C A B; D = midpoint A B

\hspace{4pt} (stage 2) A B = segment A B; C = on\_circle C A B; D = midpoint A B

\hspace{4pt} (stage 2) A B = segment A B; C = on\_circle C A B; D = midpoint A B; E = on\_circle E A B

\hspace{4pt} (stage 2) A B = segment A B; C = on\_circle C A B; D = midpoint A B; E = on\_circle E A B; F = on\_circle F A B

\hspace{4pt} (stage 2) A B = segment A B; C = on\_circle C A B; D = midpoint A B; E = on\_circle E A B; F = on\_circle F A B; G = on\_circle G A B

\hspace{4pt} (stage 2) A B = segment A B; C = on\_circle C A B; D = midpoint A B; E = on\_circle E A B; F = on\_circle F A B; G = on\_circle G A B; H = on\_circle H A B

\hspace{4pt} (stage 3) A B = segment A B; C = on\_circle C A B; D = midpoint A B; E = midpoint A C

\hspace{4pt} (stage 3) A B = segment A B; C = on\_circle C A B; D = midpoint A B; E = midpoint A C; F = on\_circle F A B

\hspace{4pt} (stage 3) A B = segment A B; C = on\_circle C A B; D = midpoint A B; E = midpoint A C; F = on\_circle F A B; G = on\_circle G A B

\hspace{4pt} (stage 3) A B = segment A B; C = on\_circle C A B; D = midpoint A B; E = midpoint A C; F = on\_circle F A B; G = on\_circle G A B; H = on\_circle H A B

\hspace{4pt} (stage 3) A B = segment A B; C = on\_circle C A B; D = midpoint A B; E = midpoint A C; F = on\_circle F A B; G = on\_circle G A B; H = on\_circle H A B; I = on\_circle I A B

\hspace{4pt} (stage 3) A B = segment A B; C = on\_circle C A B; D = midpoint A B; E = on\_circle E A B; F = on\_circle F A B; G = on\_circle G A B; H = on\_circle H A B; I = midpoint B C

\hspace{4pt} (stage 3) A B = segment A B; C = on\_circle C A B; D = midpoint A B; E = midpoint B C

\hspace{4pt} (stage 3) A B = segment A B; C = on\_circle C A B; D = midpoint A B; E = lc\_tangent E C A

\hspace{4pt} (stage 3) A B = segment A B; C = on\_circle C A B; D = midpoint A B; E = on\_circle E A B; F = on\_circle F A B; G = on\_circle G A B; H = lc\_tangent H C A

\textbf{AngleClassification}

\hspace{4pt} (stage 1) A B C = triangle A B C

\hspace{4pt} (stage 3) A B C = triangle A B C; D = midpoint B C

\hspace{4pt} (stage 3) A B C = triangle A B C; D = midpoint B C; E = midpoint A C; F = intersection\_ll F A D B E

\textbf{LengthComparison}

\hspace{4pt} (stage 1) A B C = triangle A B C

\hspace{4pt} (stage 2) A B C = triangle A B C; D = midpoint B C

\hspace{4pt} (stage 3) A B C = triangle A B C; D = midpoint A B; E = midpoint A C

\textbf{Parallel}

\hspace{4pt} (stage 1) A B C = triangle A B C; D = midpoint A B; E = midpoint A C

\hspace{4pt} (stage 1) A B C = triangle A B C; D = midpoint A B; E = midpoint A C

\hspace{4pt} (stage 1) A B C = triangle A B C; D = midpoint A B; E = midpoint A C

\hspace{4pt} (stage 2) A B C = triangle A B C; D = parallelogram A B C D

\hspace{4pt} (stage 3) A B C = triangle A B C; D = midpoint A B; E = midpoint A C; F = midpoint B C

\textbf{Perpendicular}

\hspace{4pt} (stage 1) A B C = triangle A B C; D = foot A B C

\hspace{4pt} (stage 1) A B C = r\_triangle A B C

\hspace{4pt} (stage 1) A B = segment A B; C = eq\_triangle C A B; D = eq\_triangle D A B; E = on\_circle E A B

\hspace{4pt} (stage 2) A B C = triangle A B C; D = foot A B C; E = foot C A B

\hspace{4pt} (stage 2) A B C = r\_triangle A B C; D = foot A B C

\hspace{4pt} (stage 2) A B C = triangle A B C; O = circle A B C; D = foot O A B; E = foot O C A

\hspace{4pt} (stage 3) A B C D = rectangle A B C D; E = intersection\_ll A C B D

\hspace{4pt} (stage 3) A B C = triangle A B C; O = incenter A B C; D = foot O A C; E = foot O B C; F = foot O A B

\hspace{4pt} (stage 3) A B C = r\_triangle A B C; D = foot A B C; E = foot D A B

\hspace{4pt} (stage 3) A B C = triangle A B C; D = foot A B C; E = foot C A B; F = foot B A C

\textbf{Equal}

\hspace{4pt} (stage 1) A B C = triangle A B C; D = midpoint C B

\hspace{4pt} (stage 1) A B C = triangle A B C; D = midpoint C B; O = circle O A B C

\hspace{4pt} (stage 1) A B C = triangle A B C; D = angle\_bisector B A C, on\_line D C B

\hspace{4pt} (stage 2) A B C = triangle A B C; D = midpoint A B; E = midpoint A C

\hspace{4pt} (stage 2) A B C = triangle A B C; D = midpoint A B; E = midpoint A C; O = circle O A B C

\hspace{4pt} (stage 2) A B C = triangle A B C; D = midpoint A B; E = midpoint A C

\hspace{4pt} (stage 3) A B C = triangle A B C; O = circle A B C; D = on\_circle D O C, angle\_bisector C A B
}
\captionof{figure}{\textsc{Geometry Shape Generation Code for \euclid{} training}}\label{tab:geometry_codes}
\vspace{0.5mm}
\end{GeoShapeGenBox}

\begin{figure*}[t]
  \centering
    \includegraphics[width=0.75\linewidth]{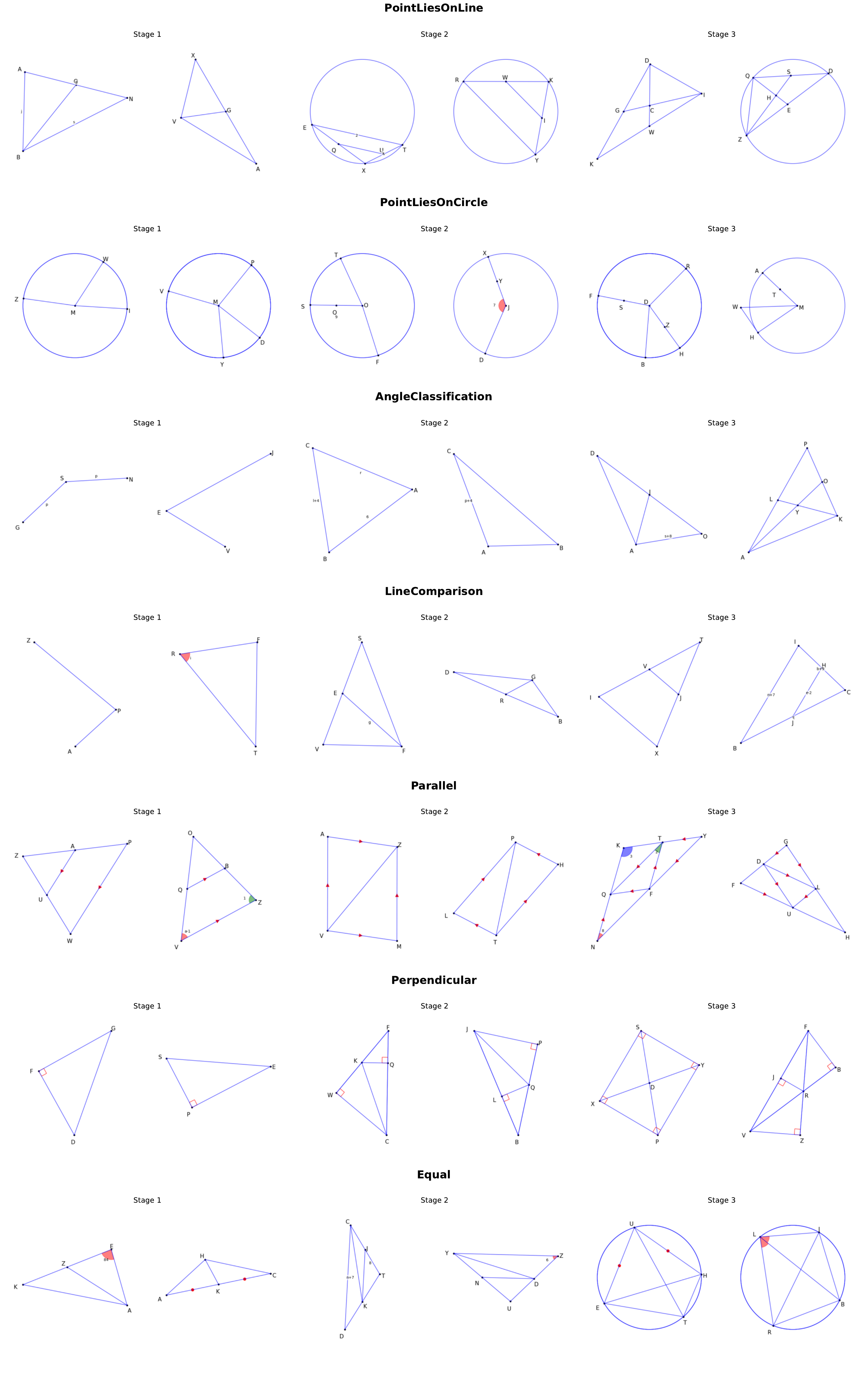}
    \caption{Examples of the geometry diagrams used to train \euclid{}, the diagrams are generated by our dataset engine.}
    \label{fig:training_shapes}
    \vspace{-3ex}
\end{figure*}

\clearpage
\section{Additional Experimental Results}

\begin{figure}[h]
  \centering
    \vspace{-1ex}
    \includegraphics[width=1\linewidth]{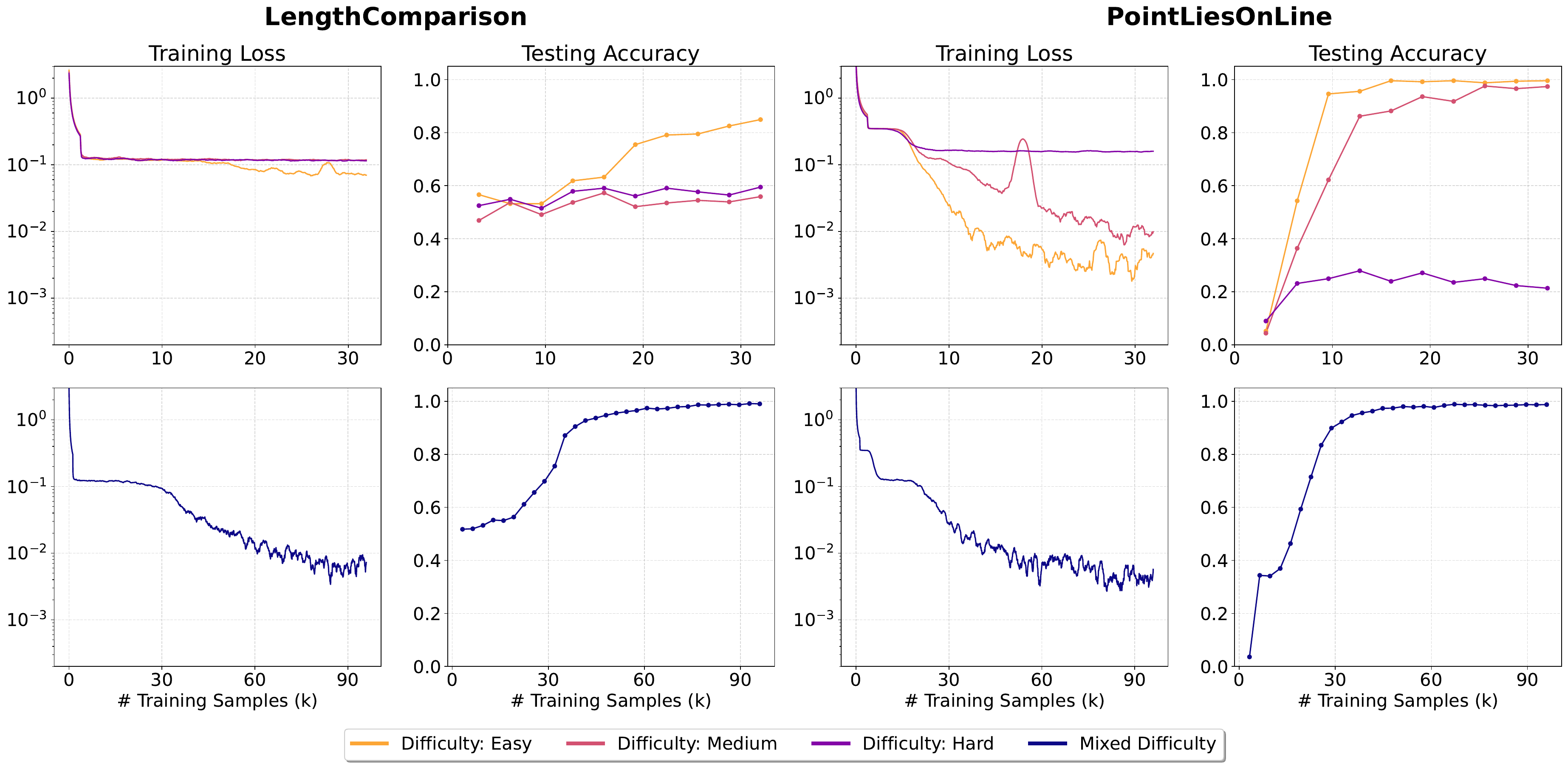}
    \vspace{-3ex}
    \caption{Result of our preliminary experiments, we use a standard setting of MLLMs: an OpenAI-CLIP@224 as visual encoders~\citep{clip}, two-layer MLP as multimodal connector and Qwen-2.5-1.5B as language model. We find that the model can reach convergence in some of the easy tasks, while struggle to learn hard tasks. We also find mixed training is better than separate training, given the same amount of training data in each difficulty level.}
    \label{appendixfig:lesson_0}
    \vspace{-1ex}
\end{figure}

\clearpage
\begin{figure}[h]
  \centering
    \vspace{-1ex}
    \includegraphics[width=1\linewidth]{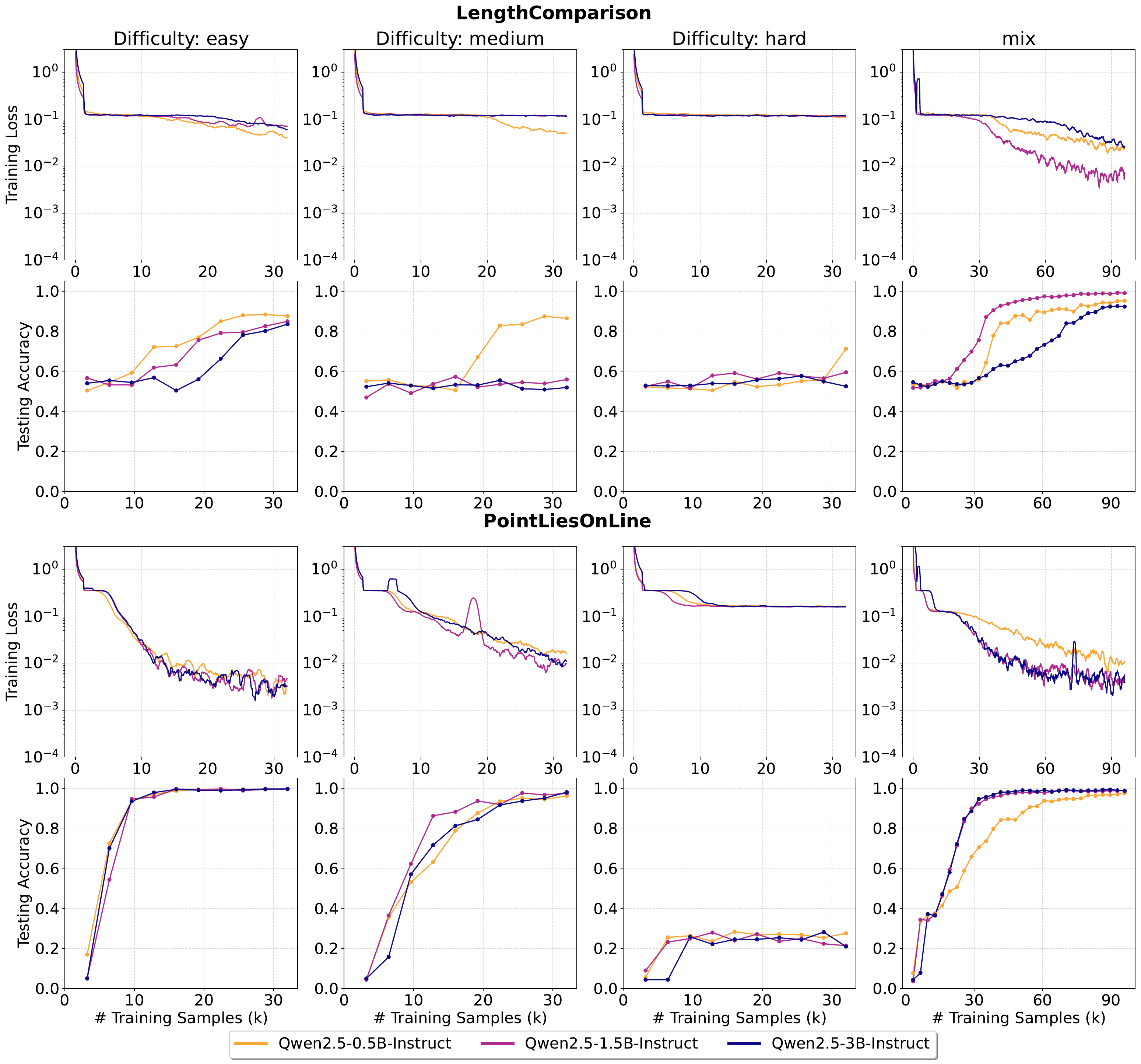}
    \vspace{-3ex}
    \caption{The complete result of the effect of LLM size. The finding is similar with~\cref{fig:llm_size}.}
    \label{appendixfig:llm_size}
    \vspace{-1ex}
\end{figure}

\clearpage
\begin{figure}[h]
  \centering
    \vspace{-1ex}
    \includegraphics[width=1\linewidth]{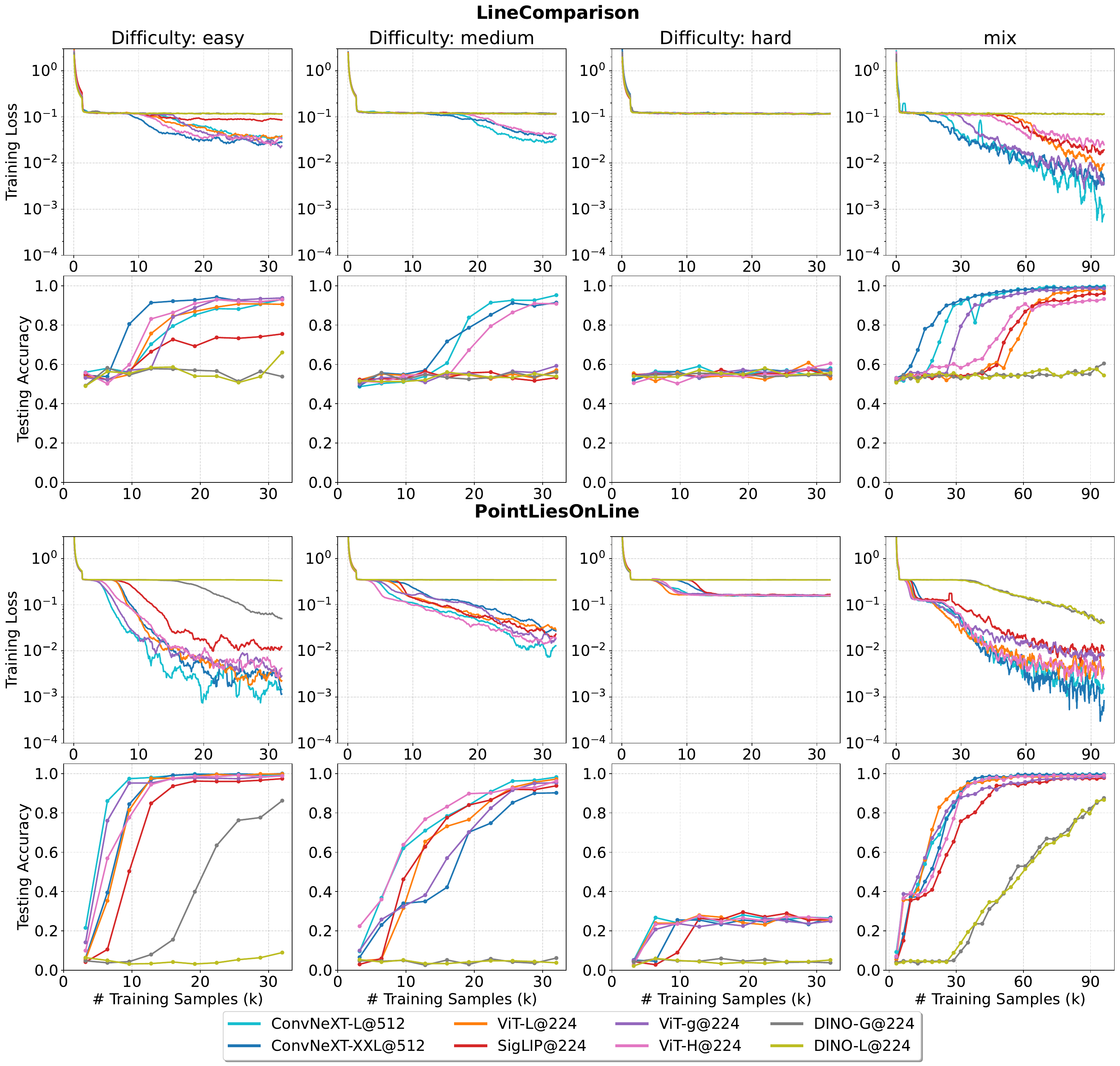}
    \vspace{-3ex}
    \caption{The complete result of the effect of different visual encoders. The finding is similar with~\cref{fig:vienc_comparison}.}
    \label{appendixfig:encoder}
    \vspace{-1ex}
\end{figure}

\clearpage
\begin{figure}[h]
  \centering
    \vspace{-1ex}
    \includegraphics[width=1\linewidth]{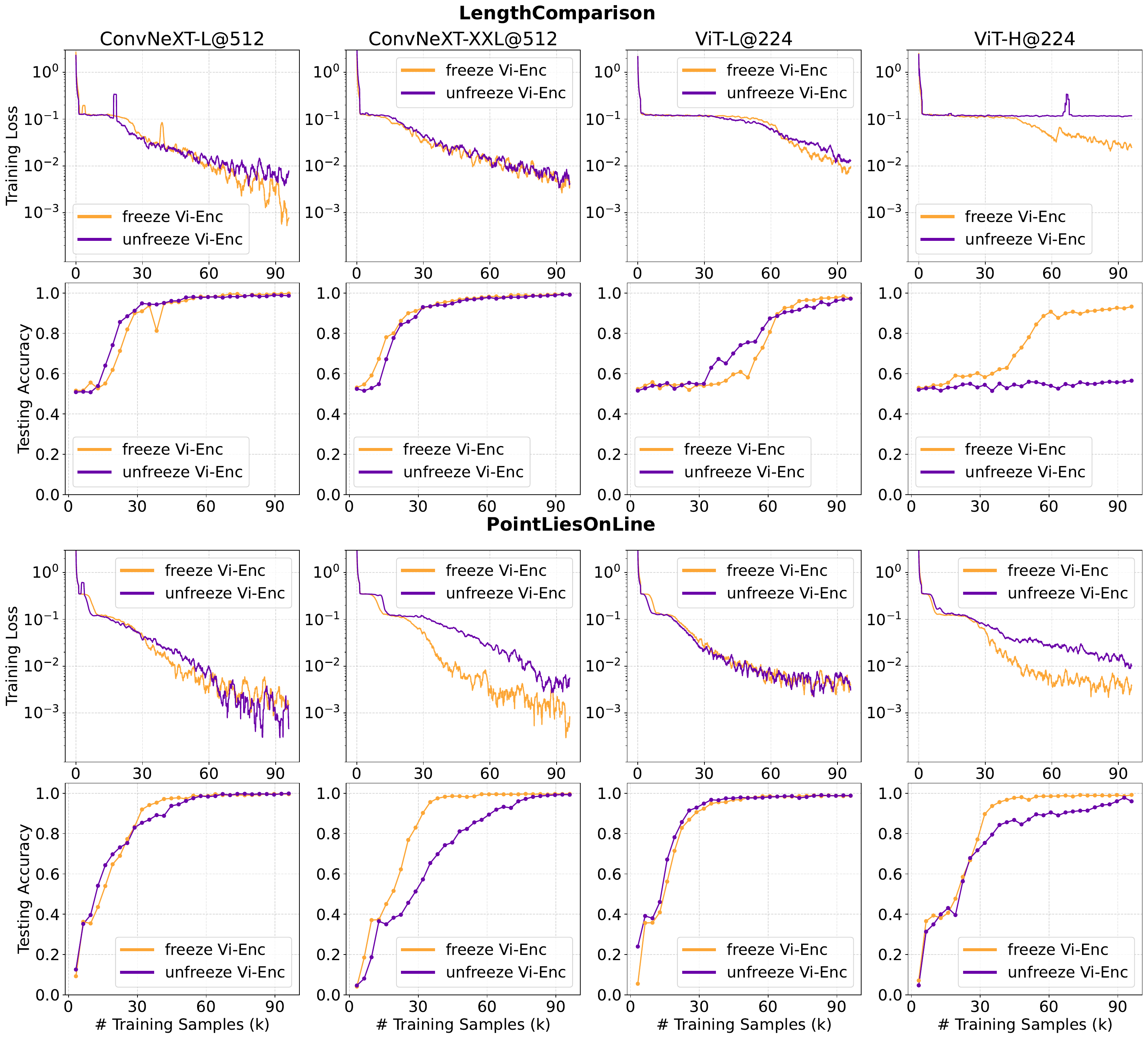}
    \vspace{-3ex}
    \caption{The complete result of the effect of tuning visual encoders. The finding is similar with~\cref{fig:freeze_unfreeze}.}
    \label{appendixfig:freeze_unfreeze}
    \vspace{-1ex}
\end{figure}

\end{document}